\documentclass[acmtog]{acmart}

\AtBeginDocument{%
  }

\usepackage{multirow}
\usepackage[capitalize]{cleveref}
\usepackage{bm}
\usepackage{enumitem}

\crefname{section}{Sec.}{Secs.}
\Crefname{section}{Section}{Sections}
\Crefname{table}{Table}{Tables}
\crefname{table}{Tab.}{Tabs.}
\newcommand{\ie}{\textit{i}.\textit{e}., }

\newcommand{\nickname}{RelightFormer}
\setcopyright{acmlicensed}
\copyrightyear{2026}
\acmYear{2026}
\setcopyright{cc}
\acmConference[SA Conference Papers '26]{SIGGRAPH Asia 2026 Conference Papers}{December 01--04, 2026}{Kuala Lumpur, Malaysia}
\acmBooktitle{SIGGRAPH Asia 2026 Conference Papers (SA Conference Papers '26), December 01--04, 2026, Kuala Lumpur, Malaysia}
\acmDOI{10.1145/3829340.3842292}
\acmISBN{979-8-4007-2842-6/2026/12}

\begin{document}
\title{\nickname{}: Feed-forward Generative Transformer for Multiview Object Relighting}

\author{Hejun Wang}
\authornote{Equal contribution and co-first authorship.}
\affiliation{%
  \institution{Shenzhen Research Institute, The Hong Kong Polytechnic University}
  \country{Hong Kong}
}
\email{hejun.wang@connect.polyu.hk}      

\author{Jinxi Li}
\authornotemark[1]
\affiliation{%
  \institution{Shenzhen Research Institute, The Hong Kong Polytechnic University}
  \country{Hong Kong}
}
\email{jinxi.li@connect.polyu.hk}

\author{Junwei Jiang}
\affiliation{%
  \institution{Shenzhen Research Institute, The Hong Kong Polytechnic University}
  \country{Hong Kong}
}
\email{junwei.jiang@connect.polyu.hk}

\author{Shiwei Mao}
\affiliation{%
  \institution{Shenzhen Research Institute, The Hong Kong Polytechnic University}
  \country{Hong Kong}
}
\email{shiwei.mao@connect.polyu.hk}

\author{Hu Cheng}
\affiliation{%
  \institution{Shenzhen Research Institute, The Hong Kong Polytechnic University}
  \country{Hong Kong}
}
\email{hucheng@polyu.edu.hk}

\author{Shouwang Huang}
\affiliation{%
  \institution{Shenzhen Research Institute, The Hong Kong Polytechnic University}
  \country{Hong Kong}
}
\email{shouwang.huang@connect.polyu.hk}

\author{Bo Yang}
\authornote{Corresponding author.}
\affiliation{%
  \institution{Shenzhen Research Institute, The Hong Kong Polytechnic University}
  \country{Hong Kong}
}
\email{bo.yang@polyu.edu.hk}

\begin{abstract}
  
Image relighting is traditionally tackled via complex inverse rendering pipelines, which suffer from ill-posed optimization, or single-image generative models that ignore crucial multi-view cues necessary for understanding 3D geometry and material interactions. To address these limitations, we introduce a feed-forward generative Transformer for direct single- and multi-view image relighting that entirely bypasses explicit intrinsic property estimation. Adapted from a video foundation model, our architecture features a latent illumination module that dynamically injects target environment maps into spatial features via cross-attention. Furthermore, we employ permutation-invariant positional encodings to symmetrically process unordered multi-view inputs without sequential bias. To train this robust data-driven model, we construct the massive Laval Objaverse Dataset (LOD), comprising 90K objects and 39K unique illuminations. Extensive experiments demonstrate state-of-the-art visual quality, photorealistic relighting quality, and strong zero-shot generalization across single-view, multi-view, and novel-view relighting tasks. Code and data for this paper are at: \textcolor{blue}{\url{https://github.com/vLAR-group/RelightFormer}}
\end{abstract}


\renewcommand{\shortauthors}{Hejun et al.}

\begin{abstract}
  
Image relighting is traditionally tackled via complex inverse rendering pipelines, which suffer from ill-posed optimization, or single-image generative models that ignore crucial multi-view cues necessary for understanding 3D geometry and material interactions. To address these limitations, we introduce a feed-forward generative Transformer for direct single- and multi-view image relighting that entirely bypasses explicit intrinsic property estimation. Adapted from a video foundation model, our architecture features a latent illumination module that dynamically injects target environment maps into spatial features via cross-attention. Furthermore, we employ permutation-invariant positional encodings to symmetrically process unordered multi-view inputs without sequential bias. To train this robust data-driven model, we construct the massive Laval Objaverse Dataset (LOD), comprising 90K objects and 39K unique illuminations. Extensive experiments demonstrate state-of-the-art visual quality, photorealistic relighting quality, and strong zero-shot generalization across single-view, multi-view, and novel-view relighting tasks. Code and data for this paper are at: \textcolor{blue}{\url{https://github.com/vLAR-group/RelightFormer}}
\end{abstract}


\begin{teaserfigure}
  \includegraphics[width=1.0\textwidth]{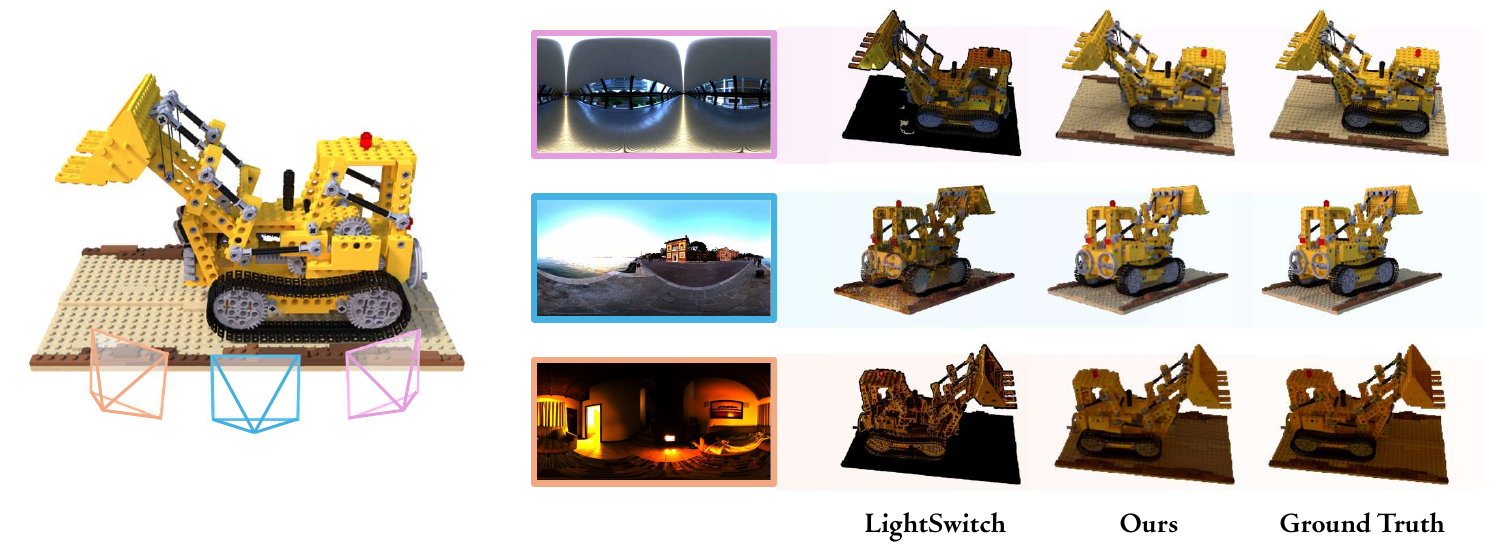}\vspace{-0.4cm}
  \caption{Given multi-view images of an object and a new illumination represented by an environment map, our \nickname{}, a feedforward generative Transformer model, relights the object with high visual fidelity.}
  \label{fig:opening}\vspace{0.2cm}
\end{teaserfigure}


\maketitle

\section{Introduction}
\label{sec:intro}
The goal of image relighting is to alter the illumination while preserving intrinsic properties such as geometry, reflectance, and overall content. This technique plays a critical role in a wide range of applications, including film production, digital content creation, gaming, and augmented reality. However, relighting is inherently challenging due to the complex interplay between light, geometry, and materials. To address this, traditional methods \cite{Kajiya1986,Debevec2000} often employ inverse rendering pipelines to estimate these intrinsic properties from input images, followed by physics-based rendering under new lighting conditions. While these approaches yield physically grounded results, they typically require sophisticated setups and rely on time-consuming per-scene optimization \cite{Zhang2021d,Zhang2021}. Furthermore, recovering intrinsic properties from 2D images is a notoriously ill-posed problem; because multiple combinations of geometry and material can produce the same visual input, incorrect estimates frequently lead to rendering artifacts under new illuminations.

Recently, to bypass the challenges of inverse rendering, several pioneering works, including Neural Gaffer \cite{Jin2024}, IllumiNeRF \cite{Zhao2024}, DiLightNet \cite{Zeng2024}, have leveraged generative models for the relighting task. Specifically, given an image of an object and a target illumination, these methods use diffusion models to directly output a relit image conditioned on the new lighting. Essentially, this direct relighting pipeline treats the outputs of the diffusion process as plausible relit results that account for various underlying combinations of intrinsic properties. Thanks to the strong priors learned from large-scale datasets, these frameworks achieve impressive object relighting results, often outperforming traditional inverse rendering pipelines. Nevertheless, these works primarily focus on single-image relighting or process multiple input images individually. Consequently, they fail to incorporate multi-view cues during training, which are crucial for accurately interpreting the underlying object geometry and lighting interactions. While the very recent method, LightSwitch \cite{Litman2025}, takes multi-view images as input, it relies on a UNet-based diffusion model that simply encodes multi-view information along the channel dimension. This approach limits the model's ability to capture rich spatial and global illumination priors, ultimately resulting in inferior relighting quality.

In this paper, we introduce \textbf{\nickname{}}, extending this direct relighting pipeline by seamlessly fusing multi-view images with a target illumination during training, while enabling the relighting of an arbitrary number of views at test time. Specifically, it builds upon Wan2.1 \cite{Wan2025}, a foundation video generation model based on flow matching \cite{Lipman2023,Esser2024}. Given images and a target illumination (\ie{} an environment map) as inputs, our method directly synthesizes the corresponding relit images without explicitly estimating any intrinsic property via inverse rendering. The core component of our approach is a newly introduced latent illumination module, which maps the target illumination into latent codes and injects them into image features via a series of cross-attention layers. Furthermore, we observe that standard video generation models inherently treat input frames sequentially due to their temporal positional encodings. However, to effectively illuminate all input images, each view should be treated equally when fusing the global environment map. To this end, we adopt a permutation-invariant positional encoding for all multi-view inputs.

Our method is built upon a feed-forward generative Transformer and standard neural layers, providing sufficient capacity for complex object relighting scenarios without relying on explicit inverse rendering. Instead, we aim to fully leverage robust visual priors learned purely from large-scale datasets, akin to the success of large foundation models in vision and language. To facilitate this data-driven approach, we propose a massive relighting dataset, namely \textbf{Laval Objaverse Dataset (LOD)}, rendered from $90{,}545$ 3D objects in Objaverse \cite{Deitke2023} and $39,008$ illuminations from Laval Database\cite{laval_indoor, laval_outdoor}. To summarize, our contributions are:

\begin{itemize}
  \vspace{-0.1cm}
    \item 
    We introduce \textbf{\nickname{}}, a large feed-forward generative Transformer for direct image relighting from single- or multi-view input, bypassing the ill-posed steps of explicit inverse rendering.
    \item To effectively adapt a video foundation model for image relighting, we introduce a latent illumination module that injects target environment maps into image features via cross-attention. Furthermore, we adopt a permutation-invariant positional encoding to eliminate sequential bias, ensuring all input views are treated symmetrically for accurate global illumination. 
    \item To fully unleash the power of data-driven visual priors, we construct a large-scale and fully open-source dataset for multi-view relighting, covering massive diverse 3D objects and abundant unique illuminations.
    \item We demonstrate state-of-the-art performance in both single-view and multi-view object relighting, outperforming existing baselines in visual quality. 
    \vspace{-0.2cm}
\end{itemize}

\section{Related Works}
\label{sec:literature}

\phantom{xW}\textbf{Inverse Rendering}: 
Inverse rendering seeks to estimate 3D geometry, material reflectance, and environment lighting from image observations, enabling downstream tasks like novel view synthesis and relighting. Traditional methods usually tackle this problem through physically-based rendering equations \cite{Kajiya1986,Debevec2000,Ramamoorthi2001,Xia2016,Nam2018,Bi2020}. Recently, with the advancement of 3D representations such as SDF \cite{Park2019}, NeRF \cite{Mildenhall2020}, and 3DGS \cite{Kerbl2023}, a series of succeeding methods have extended these representations for relightable 3D reconstruction, including methods \cite{Boss2021,Zhang2021d,Srinivasan2021,Boss2021b,Zhang2021,Hasselgren2022,Kuang2022,Boss2022,Yao2022,Zhang2022,Zhang2022b,Zhang2023,Sun2023,Jin2023,Qu2024,Tang2025} based on SDF and/or NeRF, and works \cite{Wu2024,Wei2024,Jiang2024,Du2024,Gao2024,Xu2024,Liang2024,Zhu2024,Jin2024b,Fan2025,Yong2025,Li2025,Zhang2025,Shi2025,He2025,Yao2025,Zheng2026,Han2026} based on 3DGS. While achieving impressive results, these methods typically rely on time-consuming optimization to minimize rendering losses, often incorporating hand-crafted geometry or lighting regularization terms. Consequently, these pipelines struggle to generalize to novel objects or handle challenging visual effects like specular highlights.

\textbf{Image Relighting}: Image relighting is particularly challenging, especially from a single image, because this problem is severely ill-posed. To make the problem more tractable, prior methods are often restricted to specific domains, such as portraits \cite{Debevec2000,Wenger2005,Meka2019,Zhou2019,Sun2019,Bi2021,Pandey2021,Papantoniou2023,Futschik2023,Mei2023,Ponglertnapakorn2023,Kim2024}, human bodies \cite{Tajima2021,Ji2022,Lagunas2021}, or outdoor scenes \cite{Ren2015,Li2020,Griffiths2022}. While achieving excellent results, they struggle to generalize and are often limited to specific reflection components. In contrast, our \nickname{} does not rely on restrictive assumptions or predefined lighting models; instead, it utilizes a generalizable, feed-forward Transformer trained on large-scale, diverse datasets.

\textbf{Generative Relighting}: Recently, advancements in diffusion models for image \cite{Rombach2022} and video generation \cite{Blattmann2023} have inspired an increasing number of works \cite{Jin2024,Kocsis2024,Zhao2024,Zeng2024,Phongthawee2024,Zhang2025b,Liu2025,Liang2025,He2025b,Liang2026,Dihlmann2026} to formulate relighting as a generative task. Although these methods achieve high-quality results, most focus exclusively on single-view inputs, failing to leverage the critical multi-view cues necessary for physically consistent relighting. While a handful of very recent approaches \cite{Zhang2025,Litman2025,Dihlmann2026} have begun to utilize multi-view images, they typically rely on na\"ive channel-wise concatenation within UNet-based diffusion models. This lacks a dedicated mechanism to deeply integrate multi-view information with the target illumination, ultimately leading to suboptimal results.

\section{\nickname{}}
\label{sec:method}

\subsection{Preliminaries: Latent Diffusion Transformers}
\label{subsec: dit}
Our architecture builds upon Wan2.1 \cite{Wan2025}, a powerful pre-trained latent video diffusion framework, comprising a Variational Autoencoder (VAE)~\cite{vae} for latent space compression and a Diffusion Transformer (DiT)~\cite{dit} for latent space denoising. 

The generative process adopts Rectified Flow Matching~\cite{flow}, which establishes a direct, linear trajectory between the data distribution and a standard Gaussian prior. Specifically, the forward process constructs straight-line paths via linear interpolation:
\vspace{-0.1cm}
\begin{equation} 
    \label{eq:forward}
    z_t = (1-t)z_0 + t \epsilon
\end{equation}\vspace{-0.1cm}
where $z_0$ denotes the clean latent representation, $\epsilon \sim \mathcal{N}(\bm{0}, \mathbf{I})$ is a standard Gaussian noise, and $t \in [0, 1]$ represents the continuous timestep. To reverse this process and synthesize clean images, the DiT learns a velocity field $v_\theta(z_t, t)$. Generation is then formulated as solving an Ordinary Differential Equation (ODE) that transports samples from noise distribution to data distribution:
\begin{equation} \vspace{-0.2cm}
    \label{eq:ode}
    \frac{dz_t}{dt} = v_\theta(z_t, t)
\end{equation} \vspace{-0.2cm}

Given a set of $N$ reference images, the $i^{th}$ image is denoted by $I_i^{L_0} \in \mathbb{R}^{3 \times H \times W}$ with known camera extrinsics $T_i \in \mathbb{R}^{4\times 4}$ and intrinsics $K_i \in \mathbb{R}^{3\times 3}$, and unknown original illumination $L_0$. Our goal is to synthesize a target image $I_i^{L} \in \mathbb{R}^{3 \times H \times W}$ under a new illumination condition, represented by an environment map $L \in \mathbb{R}^{3 \times H_L \times W_L}$, while preserving the original geometry and material properties. 

To achieve this, we first employ a VAE encoder $\mathcal{E}$~\cite{vae} to map each reference image $I_i^{L_0}$ into a latent representation $\mathcal{E}(I_i^{L_0})$. 
Similarly, following LuxDiT~\cite{luxdit}, we tone-map the target environment map $L$ into logarithmic and LDR versions, denoted as $L^{\text{log}} \in \mathbb{R}^{3 \times H_L \times W_L}$ and $L^{\text{ldr}} \in \mathbb{R}^{3 \times H_L \times W_L}$, respectively, to prevent excessively large values:
\begin{equation} \vspace{-0.1cm} \label{eq:tonemap}
    L^{\text{ldr}} = \frac{L}{1 + L} \cdot \left(1 + \frac{L}{M_{\text{ldr}}^2}\right), \quad\quad
    L^{\text{log}} = \frac{\log(1 + L)}{\log(1 + M_{\text{log}})}
\end{equation}
where $M_{\text{ldr}} = 16 $ and $M_{\text{log}} = 10{,}000 $, and all operations are element-wise. Both are subsequently encoded into the latent space, yielding $\mathcal{E}(L^{\text{log}})$ and $\mathcal{E}(L^{\text{ldr}})$.

To inject 3D spatial information into latent representations, we convert camera parameters into ray representations: $r_i \in \mathbb{R}^{6 \times H \times W}$ for the reference images and $r^L \in \mathbb{R}^{6 \times H_L \times W_L}$ for the environment map, parameterized by Plücker coordinates, \textit{i.e.}, ray direction and moment. A lightweight learnable network $\phi$ is then used to project these ray maps into latent embeddings which are directly added to the corresponding reference images and the target illumination latent embeddings. Lastly, these ray-enhanced latent embeddings are patchified to obtain image tokens $x^{\text{ref}} \in \mathbb{R}^{S \times C_{\text{token}}}$ and illumination tokens $x^{\text{illum}} \in \mathbb{R}^{S_{\text{illum}} \times C_{\text{token}}}$. Here, $S$ and $S_{\text{illum}}$ denote the respective sequence lengths, and $C_{\text{token}}$ is the token dimension.

\begin{figure*}[t]  
\centering
\includegraphics[width=\linewidth]{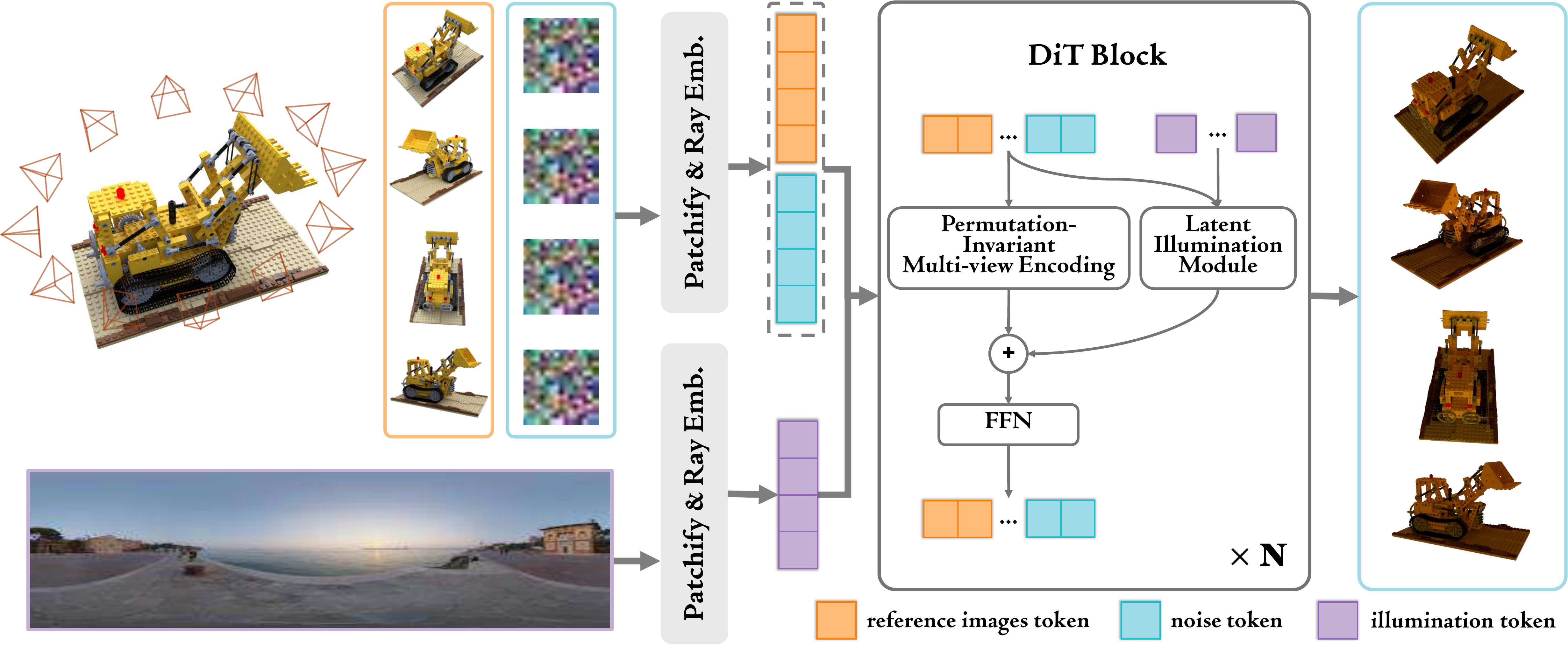} 
\vskip -0.1in
\caption{\textbf{Architecture of \nickname{}.} Input modalities (noise, reference images, and an environment map) are first patchified into token sequences, with ray embeddings added to encode stereo geometric priors. 
Noise and reference tokens are concatenated and processed through two parallel attention pathways: a multi-view self-attention module for intra- and cross-view feature aggregation, and an illumination attention module that dynamically injects lighting cues into spatial features. 
The outputs of both branches are element-wise summed and passed through an FFN. 
After the final layer, reference tokens are discarded, and the updated noise tokens are used to predict the flow-matching velocity field. 
For clarity, VAE encoding and decoding stages are omitted.}
\label{fig:method}
\vspace{-0.4cm}
\end{figure*}

As required by the generative process, we sample noise from a standard Gaussian distribution matching the shape of $\mathcal{E}(I_i^{L})$. We then apply the same ray embedding and patchification pipeline to the noisy target latent $z_t$ at timestep $t$ to obtain noise tokens $x_t^{\text{noise}}$ $\in \mathbb{R}^{S \times C_{\text{token}}}$. 
Following ReCamMaster~\cite{recammaster}, we concatenate $x^{\text{ref}}$ and $x_t^{\text{noise}}$ along the sequence dimension to form a unified image token sequence: $x_t^{\text{image}}$ $= [x^{\text{ref}}, x_t^{\text{noise}}] \in \mathbb{R}^{2S \times C_{\text{token}}}$. 
After preparing all token sequences, namely the image tokens $x_t^{\text{image}}$ and illumination tokens $x^{\text{illum}}$, we feed them into our Transformer, where they are jointly processed through a series of blocks.
Lastly, we discard the first half of the transformed image tokens (corresponding to reference views) and retain the second half to predict the velocity $v_\theta(z_t, t)$ in Equation~\ref{eq:ode}. Figure~\ref{fig:method} shows our architecture.

\subsection{Latent Illumination Module}
Physically, observed surface radiance $L_o(\omega_o)$ results from integrating incident illumination $L_i(\omega_i)$, where each directional contribution is modulated by local surface geometry and material properties, \ie{} the bidirectional scattering distribution function (BSDF) $f(\omega_i,\omega_o)$, as illustrated by the rendering equation~\cite{Kajiya1986}:
\begin{equation}
    L_o(\omega_o) = \int_{\Omega} f(\omega_i,\omega_o) \, L_i(\omega_i) \, (\omega_i \cdot n) \, \mathrm{d}\omega_i
    \label{eq:rendering_equation_}
\end{equation}
where the hemisphere $\Omega$ is centered at the surface normal $n$, and $\omega_i$ and $\omega_o$ denote the incident and outgoing directions, respectively.

Mathematically, the cross-attention mechanism computes an output feature $o_i$ as a weighted summation of value vectors $v_j$:
\begin{equation}\vspace{-0.2cm}
    o_i = \sum_j \underset{j}{\operatorname{softmax}}\!\left(\frac{q_i k_j^\top}{\sqrt{d}}\right) \, v_j,
    \label{eq:attention_break_down_}
\end{equation}
where $q_i$ and $k_j$ denote the query and key embeddings, respectively; $v_j$ is the value embedding; $d$ is their channel dimension; and the softmax is evaluated over $j$.

Motivated by this structural similarity between Eq.~\ref{eq:rendering_equation_} and Eq.~\ref{eq:attention_break_down_}, we design an illumination attention module that approximates the rendering integral in a discrete, learnable manner. Specifically, at Transformer block $\ell$, we combine multi-view self-attention and illumination cross-attention as:
\begin{equation}\vspace{-0.2cm}
    x_{t,\ell}^{\text{cond}} = \underbrace{\operatorname{GTA}\!\left(x_{t,\ell}^{\text{image}}\right)}_{\text{multi-view self-attention}} + \underbrace{\operatorname{CrossAttn}\!\left(x_{t,\ell}^{\text{image}};x^{\text{illum}}\right)}_{\text{illumination cross-attention}}
    \label{eq:illum_attn_injection}
\end{equation}
where $t$ and $\ell$ denote the diffusion timestep and block index, respectively. $\operatorname{GTA}(X)$ denotes Geometry-Aware Attention~\cite{gta} with its queries, keys, and values all projected from $X$; its detailed formulation is provided in Section~\ref{subsec:permutation_encoding}. The resulting conditioned tokens $x_{t,\ell}^{\text{cond}}$ are subsequently processed by a feed-forward network (FFN) to produce the block output. 

In the illumination attention block, cross-attention is computed between image tokens, which serve as queries $q$, and illumination tokens that encode incident lighting information $L_i$, which act as both keys $k$ and values $v$. 
In this formulation, the attention weights 
$\underset{j}{\operatorname{softmax}}\!\left(q_i k_j^\top/\sqrt{d}\right)$ implicitly learn to approximate the product $f(\omega_i,\omega_o)(\omega_i \cdot n)$, dynamically modulating how much each incident light direction contributes to the final appearance. The queries are derived from the reference and noisy target image features, while the keys are constructed from ray-embedded incident light representations, enabling the network to selectively gather relevant illumination cues for each spatial location.

\subsection{Permutation-Invariant Multi-view Encoding}
\label{subsec:permutation_encoding}
Unlike video sequences governed by temporal ordering, multi-view images constitute an inherently unordered observation set. 
Intuitively, our architecture must be permutation-invariant with respect to the input view order to guarantee robust cross-view reasoning. To this end, we replace the frame-indexed RoPE~\cite{rope} inherited from the vanilla Wan2.1~\cite{Wan2025} with PRope~\cite{prope}, which achieves permutation invariance by design: rotation angles are computed directly from physical camera configurations rather than arbitrary frame indices. 
Specifically, the vanilla self-attention mechanism is defined as:
\begin{equation}\vspace{-0.2cm}
    \mathrm{Attn}(Q, K, V) = \mathrm{softmax}\!\left(\frac{QK^\top}{\sqrt{d}}\right)V,
\end{equation}
where $Q, K, V \in \mathbb{R}^{T \times d}$, $T$ is the image-token sequence length, and $d$ is the attention-head dimension. 
The multi-view image attention module adopts the Geometry-Aware Attention (GTA) mechanism~\cite{gta}:
\begin{equation}\vspace{-0.2cm}
    \mathrm{GTA}(Q, K, V) = D  \mathrm{Attn}\!\big( D^\top Q,\; D^{-1} K,\; D^{-1} V \big),
\end{equation}
where $D \in \mathbb{R}^{T \times d \times d}$ is a sequence of per-token transformation matrices derived from camera projection matrices and 2D patch coordinates, with multiplication by $D$ applied token-wise. 
Unlike standard RoPE~\cite{rope}, which only encodes 2D grid positions, PRope~\cite{prope} decomposes each token's transformation matrix $D_i$ into projective and positional components:
\begin{align} 
    D_i &= \begin{bmatrix} D_i^{\text{Proj}} & \mathbf{0} \\ \mathbf{0} & D_i^{\text{RoPE}} \end{bmatrix}, 
    \quad
    D_i^{\text{Proj}} = I_{d/8} \otimes \tilde{P}_{i} \in \mathbb{R}^{\frac{d}{2} \times \frac{d}{2}}, \label{eq:d_proj} \\
    D_i^{\text{RoPE}} &= \begin{bmatrix} \operatorname{RoPE}_{d/4}(x_i) & \mathbf{0} \\ \mathbf{0} & \operatorname{RoPE}_{d/4}(y_i) \end{bmatrix} \in \mathbb{R}^{\frac{d}{2} \times \frac{d}{2}}, \label{eq:d_rope}
\end{align}
where $D_i^{\text{Proj}}$ encodes the relative projective geometry between camera frustums via the normalized projection matrix $\tilde{P}_{i} \in \mathbb{R}^{4 \times 4}$, and $D_i^{\text{RoPE}}$ applies standard rotary embeddings to the patch coordinates $(x_i, y_i)$; $I_{d/8}$ is the identity matrix and $\otimes$ denotes the Kronecker product. Consequently, the multi-view attention module achieves permutation invariance, as the positional encoding depends solely on the underlying camera configuration and spatial layout, regardless of the input token order. More details concerning our methodology are provided in Appendix \ref{app:method}.

\section{Laval Objaverse Dataset}
\label{sec:dataset}

Due to the lack of public large-scale datasets for multi-view object relighting, we build \textbf{Laval Objaverse Dataset (LOD)} comprising diverse 3D objects under abundant lighting conditions.

Specifically, we render multi-view images from \textbf{90,545 high-quality objects} from Objaverse\cite{Deitke2023}, curated by Neural Gaffer~\cite{Jin2024}. For illumination, we leverage the Laval Indoor HDR Dataset~\cite{laval_indoor} and Laval Outdoor HDR Dataset~\cite{laval_outdoor} as our environment map sources. 
To further enhance diversity, we augment each HDR map with uniformly sampled horizontal rotations, expanding the pool by 16$\times$. This yields \textbf{39,008 unique illumination conditions} for comprehensive coverage.
For each object, we randomly sample 16 environment maps (8 indoor + 8 outdoor) and use Cycles~\cite{cycle} to render 16 randomly sampled views for training and 200 views for validation/testing per object-lighting pair. Critically, camera viewpoints vary across objects, preventing the model from overfitting to fixed view configurations. To ensure rigorous evaluation, we enforce strict splits across objects, illumination conditions, and camera viewpoints to prevent data leakage.
More details concerning Laval Objaverse Dataset are provided in Appendix \ref{app:dataset}.

\section{Experiments}
\label{sec:exp}
\label{sec:experiment}

We fine-tune \nickname{} on our Laval Objaverse Dataset. 
To support flexible input configurations, we randomly sample $1$--$16$ reference images per iteration to train a single large model. 
Our optimization trains at a resolution of $256 \times 256$ for 80K steps across 4 NVIDIA H200 GPUs, with a global batch size of 128 and an initial learning rate of $1 \times 10^{-4}$ decayed via cosine annealing. 

\subsection{Single Image Relighting}\label{subsec:single_image_relighting}

\begin{table*}[t!]
    \centering
    \scriptsize
    \caption{Quantitative results of image relighting on our dataset under different input view settings.} \vspace{-0.2cm}
    \label{tab:merged_relighting}
    \setlength{\tabcolsep}{3.5pt} 
    \resizebox{\textwidth}{!}{
    \begin{tabular}{l cccc cccc cccc}
        \toprule
        \multirow{2}{*}{} &
        \multicolumn{4}{c}{\textbf{Single View}} &
        \multicolumn{4}{c}{\textbf{16 Views}} &
        \multicolumn{4}{c}{\textbf{32 Views}} \\
        \cmidrule(lr){2-5} \cmidrule(lr){6-9} \cmidrule(lr){10-13}
        & sPSNR$\uparrow$ & PSNR$\uparrow$ & SSIM$\uparrow$ & LPIPS$\downarrow$ &
          sPSNR$\uparrow$ & PSNR$\uparrow$ & SSIM$\uparrow$ & LPIPS$\downarrow$ &
          sPSNR$\uparrow$ & PSNR$\uparrow$ & SSIM$\uparrow$ & LPIPS$\downarrow$ \\
        \midrule
        DiLightNet~\cite{Zeng2024}            
        & 17.83 & 15.95 & 0.777 & 0.232 
        & 16.10 & 15.00 & 0.733 & 0.316
        & 16.32 & 15.06 & 0.748 & 0.298 \\
        Neural Gaffer~\cite{Jin2024}       
        & 20.79 & 17.78 & 0.863 & 0.116 
        & 20.80 & 17.85 & 0.864 & 0.116
        & 20.81 & 17.78 & 0.857 & 0.122 \\
        Neural Gaffer$_{finetuned}$~\cite{Jin2024} 
        & 22.27 & 20.25 & 0.883 & 0.095
        & 22.24 & 20.26 & 0.884 & 0.096 
        & 22.34 & 20.26 & 0.879 & 0.099 \\
        LightSwitch~\cite{Litman2025}           
        & 18.63 & 16.56 & 0.778 & 0.235
        & 18.53 & 16.52 & 0.786 & 0.220
        & 18.48 & 16.49 & 0.787 & 0.219 \\
        Reli3D~\cite{Dihlmann2026}  
        & 18.52 & 15.54 & 0.814 & 0.254 
        & 18.65 & 15.70 & 0.817 & 0.248
        & 18.76 & 15.75 & 0.816 & 0.247 \\
        \midrule
        \textbf{\nickname{} (Ours)}         
       & \textbf{23.80} & \textbf{21.16} & \textbf{0.894} & \textbf{0.112}
        & \textbf{24.83} & \textbf{22.62} & \textbf{0.905} & \textbf{0.084}  
        & \textbf{25.07} & \textbf{22.83} & \textbf{0.906} & \textbf{0.080} \\
        \bottomrule

    \end{tabular}}
\end{table*}

\phantom{xW}\textbf{Task}: 
In testing, single-image relighting aims to modify the appearance of a given image according to a target illumination condition, while preserving the geometry, materials, and scene identity.

\textbf{Evaluation Protocol}: 
For quantitative evaluation, we curate $7{,}248$ image pairs from our held-out test set for each sub-task, uniformly sampling across all objects and illumination conditions to ensure comprehensive coverage. We report four standard metrics: \textbf{PSNR}, \textbf{SSIM}, \textbf{LPIPS}, and \textbf{scale-invariant PSNR (sPSNR)} to assess foreground quality. Notably, sPSNR aligns predicted intensities to the ground-truth scale via least-squares regression prior to error computation, thereby eliminating bias introduced by global intensity or exposure mismatches.

 \textbf{Baselines}: We compare \nickname{} against state-of-the-art methods spanning two paradigms: inverse-rendering approaches (\textit{e.g.}, LightSwitch~\cite{Litman2025}, Reli3D~\cite{Dihlmann2026}) and generative relighting frameworks (\textit{e.g.}, DilightNet~\cite{Zeng2024}, Neural Gaffer~\cite{Jin2024}). Note that LightSwitch~\cite{Litman2025} and Reli3D~\cite{Dihlmann2026} are originally designed for multi-view consistent relighting; we additionally validate their capability on the more challenging task of single-view relighting. 
For fairness, we fine-tune Neural Gaffer~\cite{Jin2024} on our training dataset until convergence and report these adapted results alongside those obtained using the original pre-trained weights.

\begin{figure*}[h!]  
\centering
\includegraphics[width=\linewidth]{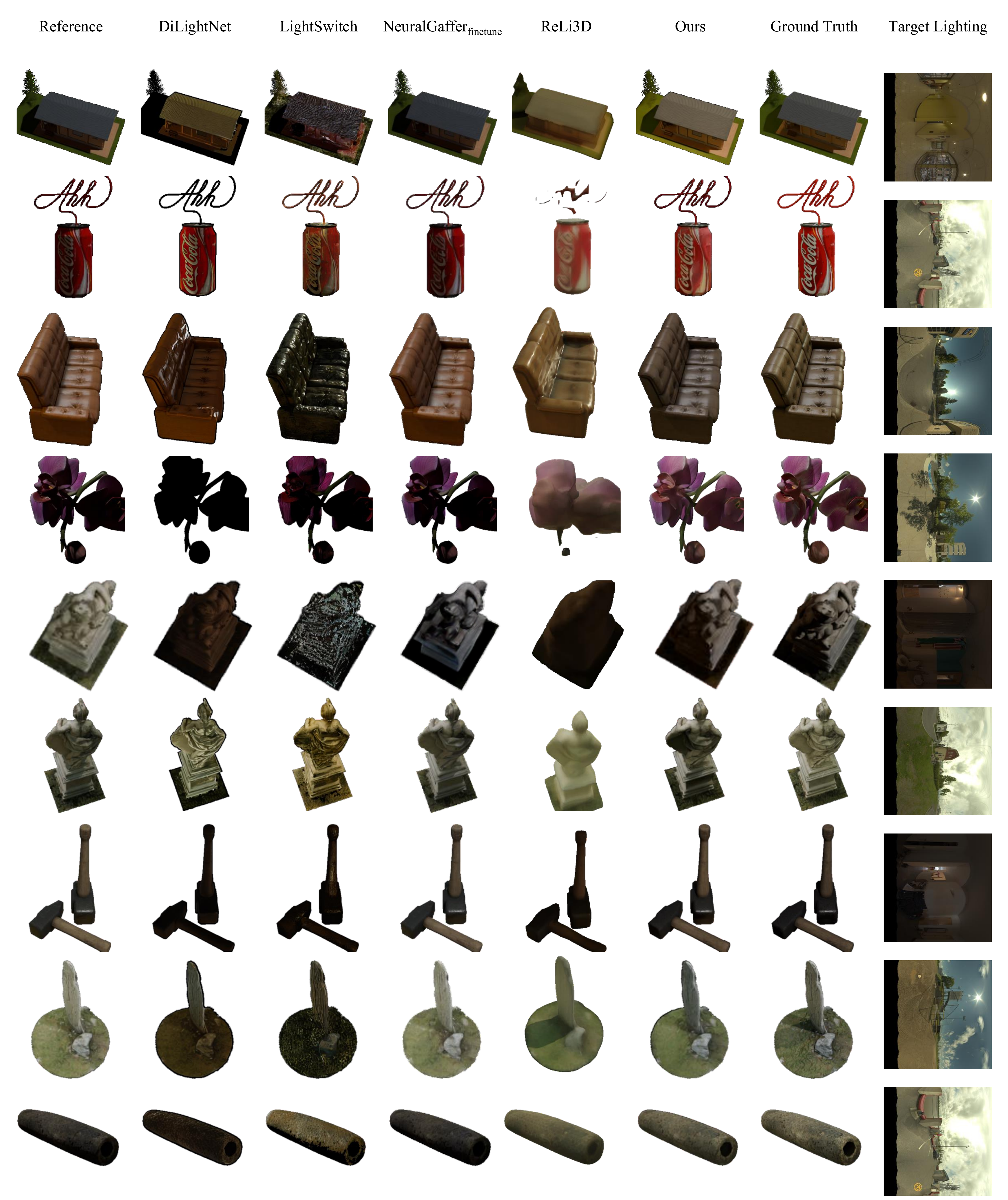} 
\caption{Qualitative results for single image relighting.}
\label{fig:qualitative_of_singl_view}
\end{figure*}

\textbf{Results and Analysis}: 
Quantitative evaluations on our dataset are presented in the left part of Table~\ref{tab:merged_relighting}. 
Our approach demonstrates superior generalizability to unseen objects, maintaining consistent textures and preserving intricate details. 
Single-image relighting is inherently challenging due to the absence of multi-view correspondences, which are typically required for robust geometry and material estimation. 
Consequently, LightSwitch~\cite{Litman2025} and Reli3D~\cite{Dihlmann2026} suffer from performance degradation in ambiguous cases, as illustrated in Figure~\ref{fig:qualitative_of_singl_view}.
In contrast, our model exhibits strong robustness to such ambiguities, generating photorealistic results by leveraging priors learned from our large-scale training dataset. \vspace{-0.2cm}

\subsection{Multi-view Image Relighting}\label{subsec:multi_image_relighting}

\phantom{xW}\textbf{Task}: 
During testing, multi-view image relighting aims to synthesize two or more target images under a specified illumination condition, conditioned on reference images captured from viewpoints identical to the target images. 

\textbf{Evaluation Protocol}:
We define two evaluation sub-tasks to assess scalability: $16$-to-$16$ (multi-view), and $32$-to-$32$ (dense multi-view) relighting, each comprising $7{,}248$ multi-view image pairs similarly. To further validate generalization, we additionally evaluate on 484 pairs from TensoIR dataset~\cite{Jin2023}, where each pair comprises 16 reference images captured from distinct viewpoints.

\textbf{Baselines}: Similarly, we compare our \nickname{} model against DilightNet \cite{Zeng2024}, LightSwitch \cite{Litman2025}, Reli3D \cite{Dihlmann2026}, Neural Gaffer \cite{Jin2024}, along with our finetuned Neural Gaffer \cite{Jin2024} and report the same metrics. For single-view methods, DilightNet \cite{Zeng2024} and Neural Gaffer \cite{Jin2024}, which inherently accept only one reference image as input, we evaluate their multi-view capability by processing each target view independently using its corresponding reference, then aggregating results across all views. 

\begin{figure*}[th!]  
\centering
\includegraphics[width=\linewidth]{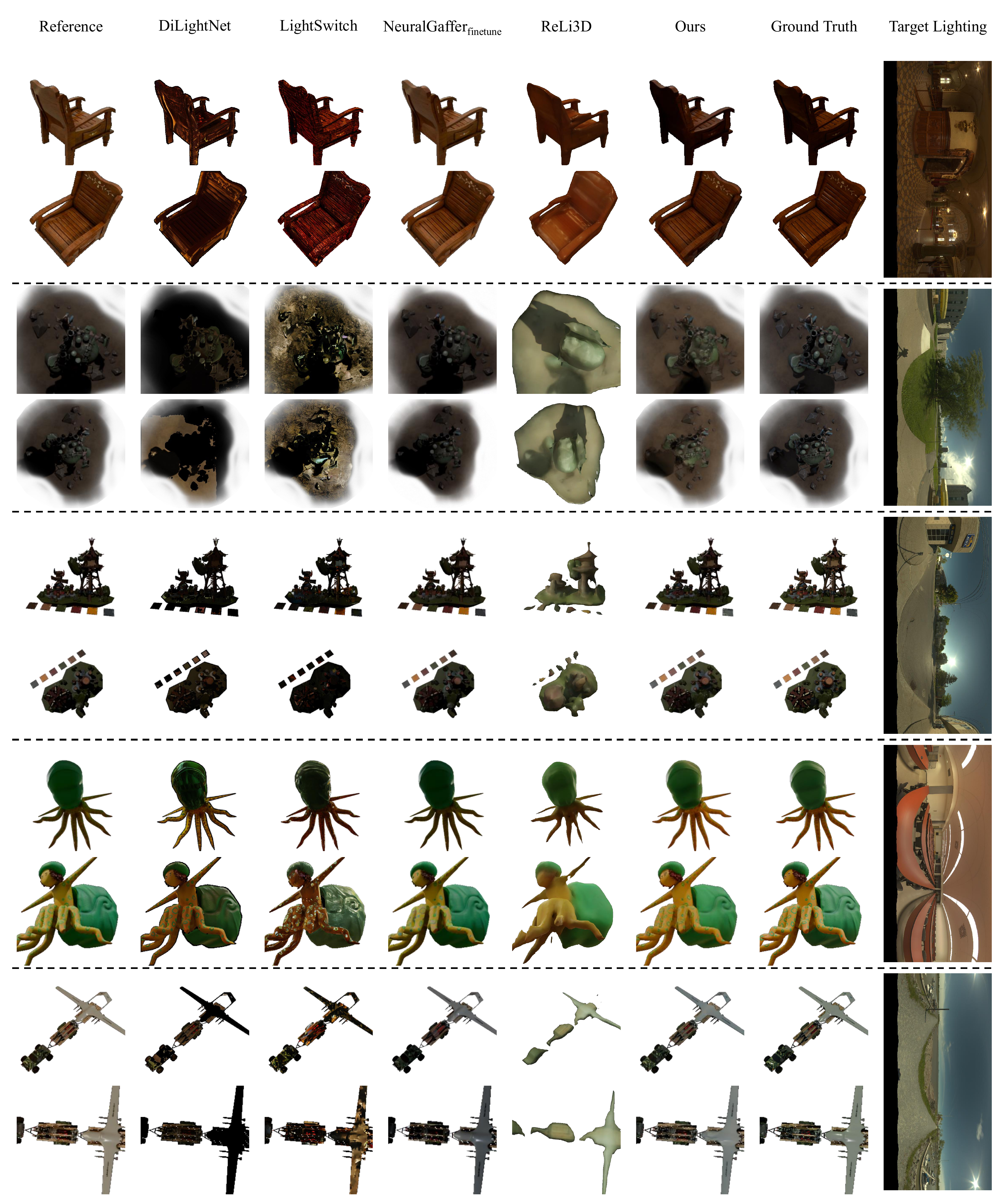} 
\vskip -0.1in
\caption{More qualitative results for multi-view image relighting.}
\label{fig:multi_view_qualitative}
\end{figure*}

\textbf{Results and Analysis}: We report quantitative results on Table \ref{tab:merged_relighting}.
Figure \ref{fig:multi_view_qualitative} showcases some qualitative examples for multi-view image relighting.
\nickname{} consistently outperforms all baselines, demonstrating that our architecture more effectively integrates illumination conditions and geometric priors. 
Notably, performance improves monotonically with an increasing number of reference views, indicating that our model successfully extracts underlying 3D geometry and harmoniously fuses it with global incident lighting. 
However, DilightNet\cite{Zeng2024} and Neural Gaffer\cite{Jin2024} cannot distill cues from multi-view correspondence, resulting in low-fidelity performance. Furthermore, \nickname{} achieves competitive results on the TensoIR dataset~\cite{Jin2023}, which lies out-of-distribution relative to our training data. Please refer to Appendix ~\ref{app:experiment} for more details. 

\subsection{Real-world Generalization and Material Analysis}
\label{subsec:real_world_material}

\phantom{xW}\textbf{Task}: We further evaluate zero-shot generalization on the 42-object validation set of the real-world OLATverse dataset~\cite{Zhou2026b}. We compare with DiLightNet~\cite{Zeng2024}, Neural Gaffer~\cite{Jin2024} and LightSwitch~\cite{Litman2025} on environment-map / point-light relighting. 
For environment-map relighting, we compose the OLAT images captured under one set of lights into the source environment map and use another composition as the target illumination. To examine view-dependent effects under spatially concentrated illumination, we additionally conduct rotating point-light relighting by randomly sampling two point-light maps as the source and target illumination for each object.

\textbf{Results and Analysis}: As shown in Table~\ref{tab:olatverse_realworld}, \nickname{} achieves comparative results across all metrics in environment-map relighting, demonstrating strong transfer from synthetic training data to real objects and illumination. In the more challenging rotating point-light relighting experiment, the concentrated light source produces pronounced, view-dependent highlights and shadows as its direction changes. We achieve the best PSNR and SSIM scores in this setting, while Neural Gaffer obtains slightly better sPSNR and LPIPS. These complementary metrics indicate that \nickname{} preserves overall radiometric fidelity well, although perceptual details under rapidly varying specular illumination remain challenging.

\begin{table}[t!]
    \centering
    \scriptsize
    \setlength{\tabcolsep}{2.0pt}
    \vspace{-0.1cm}
    \caption{Zero-shot evaluation on the real-world OLATverse dataset under environment-map relighting and rotating point-light relighting.} \vspace{-0.2cm}
    \label{tab:olatverse_realworld}
    \resizebox{\linewidth}{!}{
    \begin{tabular}{lcccccccc}
        \toprule
        & \multicolumn{4}{c}{Environment-map relighting} & \multicolumn{4}{c}{Rotating point-light relighting} \\
        \cmidrule(lr){2-5}\cmidrule(lr){6-9}
        Method & sPSNR$\uparrow$ & PSNR$\uparrow$ & SSIM$\uparrow$ & LPIPS$\downarrow$ & sPSNR$\uparrow$ & PSNR$\uparrow$ & SSIM$\uparrow$ & LPIPS$\downarrow$ \\
        \midrule
        DiLightNet~\cite{Zeng2024}
        & 15.06 & 16.32 & 0.748 & 0.298
        & 30.59 & \underline{22.83} & 0.718 & 0.396 \\
        
        Neural Gaffer~\cite{Jin2024} 
        & 18.56 & 15.98 & 0.958 & \underline{0.048}
        & 21.05 
        & 19.28 & \underline{0.900} & \textbf{0.122} \\
        
       Neural Gaffer$_{finetuned}$~\cite{Jin2024} 
       & \underline{19.61} & \underline{16.64} & \underline{0.961} & 0.049
       & \underline{30.85} & 11.59 & 0.040 & 0.527 \\
       
        LightSwitch~\cite{Litman2025} 
        & 12.22 & 9.88 & 0.917 & 0.144 
        & 21.93 & 15.58 & 0.886 & 0.236 \\
        
        \textbf{\nickname{} (Ours)} 
        & \textbf{20.48} & \textbf{17.23} & \textbf{0.964} & \textbf{0.047} 
        & \textbf{31.16} & \textbf{29.39} & \textbf{0.826} & \underline{0.209} \\
        \bottomrule
    \end{tabular}}
    \vspace{-0.5cm}
\end{table}

\begin{figure*}[!htbp]  
\centering
\includegraphics[width=\linewidth]{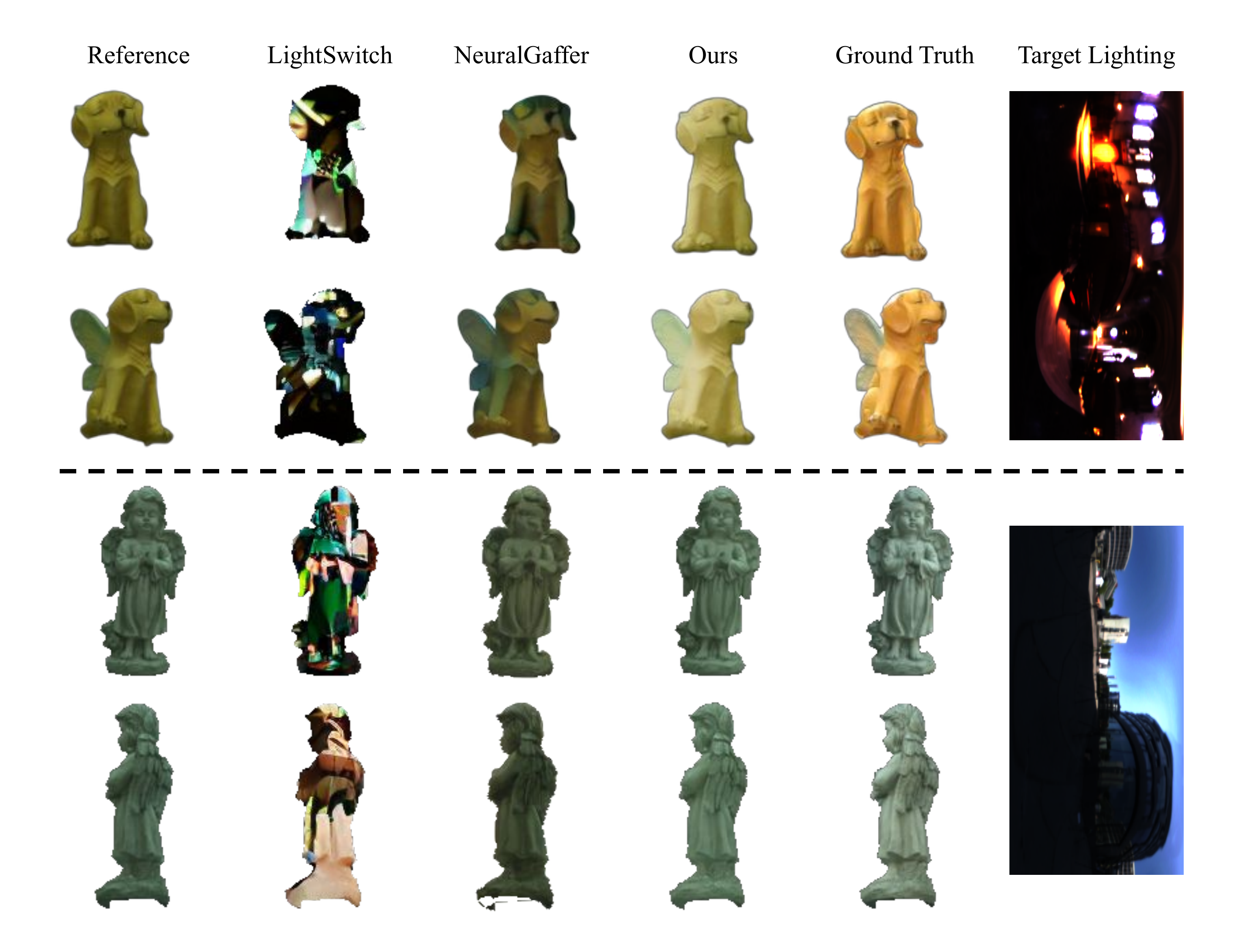} 
\vskip -0.1in
\caption{Qualitative results for multi-view image relighting in real-world OLATverse Dataset.}
\label{fig:olatverse}
\vspace{-0.4cm}
\end{figure*}

We further stratify the objects using their provided primary material labels and manually group the challenging glossy, translucent, and furry cases. 
Table \ref{tab:olatverse_materials_mask} reports PSNR/sPSNR for all material categories. \nickname{} performs best on most materials, including glossy, translucent, and furry objects. Nevertheless, we also observe that Neural Gaffer achieves a higher PSNR on plastic and stone, as image-based relighting is inherently ill-posed and ambiguous for challenging materials. Figure~\ref{fig:olatverse} shows qualitative results.

\begin{table*}[t!]
    \centering
    \scriptsize
    \caption{Material-stratified masked PSNR/sPSNR results on the real-world OLATverse dataset.}
    \vspace{-0.2cm}
    \label{tab:olatverse_materials_mask}
    \resizebox{\textwidth}{!}{%
    \begin{tabular}{lcccccccc}
        \toprule
        Method & Ceramics & Fabric & Food & Leather & Metal & Paper & Plant & Plaster \\
        \midrule
        DiLightNet~\cite{Zeng2024} & 12.02/13.18 & 10.66/11.81 & 12.51/13.49 & 11.65/12.77 & 14.26/15.25 & 11.04/13.31 & 12.18/12.25 & 11.39/12.02 \\
        Neural Gaffer~\cite{Jin2024} & 16.38/18.24 & 14.50/17.56 & 17.22/19.41 & 16.15/18.79 & 18.09/20.40 & \underline{15.10}/\underline{18.10} & \textbf{18.37}/20.02 & 15.68/18.07 \\
        Neural Gaffer$_{finetuned}$~\cite{Jin2024}  & \underline{16.88}/\underline{19.64} & \underline{16.03}/\underline{19.12} & \underline{17.64}/\underline{20.52} & \underline{17.40}/\underline{20.39} & \underline{18.40}/\underline{21.17} & 14.85/18.08 & 18.34/\underline{20.69} & \underline{16.26}/\underline{18.91} \\
        LightSwitch~\cite{Litman2025} & 10.10/12.70 & 8.90/11.17 & 9.98/12.73 & 9.48/12.54 & 10.79/14.59 & 8.76/10.37 & 11.52/14.26 & 9.87/12.21 \\
        \textbf{\nickname{} (ours)} & \textbf{17.69}/\textbf{20.00} & \textbf{16.61}/\textbf{20.01} & \textbf{18.26}/\textbf{21.18} & \textbf{17.71}/\textbf{21.09} & \textbf{19.88}/\textbf{22.84} & \textbf{16.16}/\textbf{19.40} & \underline{18.35}/\textbf{21.07} & \textbf{16.33}/\textbf{19.59} \\
        
        \midrule
        Method & Plastic & Rubber & Stone & Wax & Wood & Furry & Glossy & Translucent \\
        \midrule
        DiLightNet~\cite{Zeng2024} & 8.32/9.03 & 13.99/15.00 & 9.38/9.61 & 9.50/9.81 & 12.52/14.03 & 8.67/10.29 & 11.52/12.43 & 8.10/8.49 \\
        Neural Gaffer~\cite{Jin2024} & 14.19/16.92 & \underline{19.31}/\underline{21.66} & 13.90/16.51 & 15.04/17.99 & 16.67/19.34 & 12.05/15.82 & 15.64/17.38 & \textbf{14.06}/\underline{17.42} \\
        Neural Gaffer$_{finetuned}$~\cite{Jin2024}  & \textbf{15.42}/\underline{18.59} & 18.56/21.25 & \textbf{15.18}/\underline{17.82} & \underline{15.69}/\underline{19.10} & \underline{16.95}/\underline{20.10} & \underline{13.99}/\underline{17.43} & \underline{16.95}/\underline{19.62} & 12.99/17.03 \\
        LightSwitch~\cite{Litman2025} & 8.46/9.83 & 12.07/16.08 & 8.42/10.99 & 9.71/11.25 & 10.51/13.15 & 7.89/8.98 & 9.83/12.41 & 8.65/9.74 \\
        \textbf{\nickname{} (ours)} & \underline{15.27}/\textbf{18.69} & \textbf{20.34}/\textbf{23.84} & \underline{15.06}/\textbf{18.09} & \textbf{15.78}/\textbf{19.97} & \textbf{17.79}/\textbf{21.40} & \textbf{14.28}/\textbf{17.75} & \textbf{17.40}/\textbf{19.91} & \underline{13.61}/\textbf{18.51} \\
        
        \bottomrule
    \end{tabular}%
    }
    \vspace{-0.1cm}
\end{table*}

\subsection{Novel View Relighting}
\label{subsec:novel_view_relighting}

\phantom{xW} \textbf{Task}: 
Novel view relighting aims to synthesize photorealistic images of a scene from unseen camera viewpoints under target illumination conditions, given one or more input images. This capability offers greater flexibility and broader applicability compared to single- and multi-view image relighting, as it enables free-viewpoint navigation while preserving consistent relighting and geometry. Our model is primarily designed for input-view relighting; novel-view relighting additionally requires reconstructing unseen viewpoints.

\textbf{Post-training on Novel View Supervision}: 
Since our original training scheme does not explicitly supervise novel viewpoints, we introduce a novel-view post-training phase. This phase optimizes for $20$K steps with a fixed learning rate of $1 \times 10^{-4}$. During post-training, our model takes a variable number of reference images ($1$--$16$) as input, while dense supervision is consistently applied across $16$ target viewpoints. 
This additional stage yields a specialized variant, denoted as \nickname-Post.
\begin{table}[htbp]\vspace{-0.2cm}
    \centering
    \centering
    \setlength{\tabcolsep}{1pt}
    \setlength{\intextsep}{0pt}
    \setlength{\belowbottomsep}{1pt}
    \caption{Quantitative comparison on Stanford-ORB benchmark.}\vspace{-0.3cm}
    \label{tab:novel_view_relighting}
    \resizebox{\linewidth}{!}{
    \begin{tabular}{llccccc}
        \toprule
        \textbf{Paradigm} & \textbf{Method} & \textbf{Time} & \textbf{PSNR-H}$\uparrow$ & \textbf{PSNR-L}$\uparrow$ & \textbf{SSIM}$\uparrow$ & \textbf{LPIPS}$\downarrow$ \\
        \midrule
        \multirow{2}{*}{\textbf{Feed Forward}} 
        & Reli3D~\cite{Dihlmann2026} 
            & $\sim$2 min & 20.26 & 25.13 & 0.943 & 0.077 \\
        & \textbf{\nickname-Post (Ours)} 
            & $\sim$2 min & \textbf{20.99} & \textbf{26.84} & \textbf{0.952} & \textbf{0.061} \\
        \midrule
        \multirow{7}{*}{\textbf{Optimization}} 
        & Neural-PBIR~\cite{Sun2023} & $\sim$1 h & 26.01 & 33.26 & 0.979 & 0.023 \\
        & NVDIFFREC~\cite{nvdiffrec} & $\sim$1 h & 22.91 & 29.72 & 0.963 & 0.039 \\
        & NVDIFFREC-MC~\cite{Hasselgren2022} & $\sim$2 h & 24.43 & 31.60 & 0.972 & 0.036 \\
        & InvRender~\cite{Zhang2022} & 1--2 h & 23.76 & 30.83 & 0.970 & 0.046 \\
        & NeRFactor~\cite{Zhang2021d} & $\sim$5 h & 23.54 & 30.38 & 0.969 & 0.048 \\
        & PhySG~\cite{Zhang2021} & 10--20 h & 21.81 & 28.11 & 0.960 & 0.055 \\
        & \textbf{3DGS+\nickname{} (Ours)} & $\sim$10 min & 23.25 & 30.12 & 0.967 & 0.038 \\
        \bottomrule
    \end{tabular}}
    \vspace{-0.2cm}
\end{table}

\textbf{Validation on Real-world Dataset}: 
We compare \nickname-Post against the state-of-the-art 3D reconstruction based baseline Reli3D \cite{Dihlmann2026} in a zero-shot setting on the real-world Stanford-ORB benchmark~\cite{kuang2023stanfordorb}. Since Stanford-ORB provides ground truth only for novel-view relighting, we evaluate our method either through novel-view post-training (\nickname-Post) or by reconstructing novel views with 3DGS before applying our original relighting model. We also include optimization-based inverse-rendering methods reported by the benchmark. Note that Reli3D is trained with ground truth 3D object shapes for disentangling illumination, enabling it to be extremely powerful in relighting.  

Table~\ref{tab:novel_view_relighting} shows quantitative results where
\nickname-Post achieves on par with, or even slightly better performance than Reli3D across all metrics, at comparable inference time.
Optimization-based methods obtain higher reconstruction quality by optimizing decomposed intrinsic properties for each scene, but require one to twenty hours. In contrast, our latent-space pipeline completes inference within minutes, while 3DGS+\nickname{} offers an intermediate quality--efficiency trade-off. Moreover, the latent-space formulation can naturally accommodate additional conditioning modalities, such as text. We therefore view our approach as complementary to, rather than a replacement for, optimization-based inverse-rendering pipelines.
More details regarding our experiment are provided in Appendix ~\ref{app:experiment}.

\section{Ablation Analysis}
\label{sec:anlysis}
\begin{figure*}[t!]  
\centering
\includegraphics[width=\linewidth]{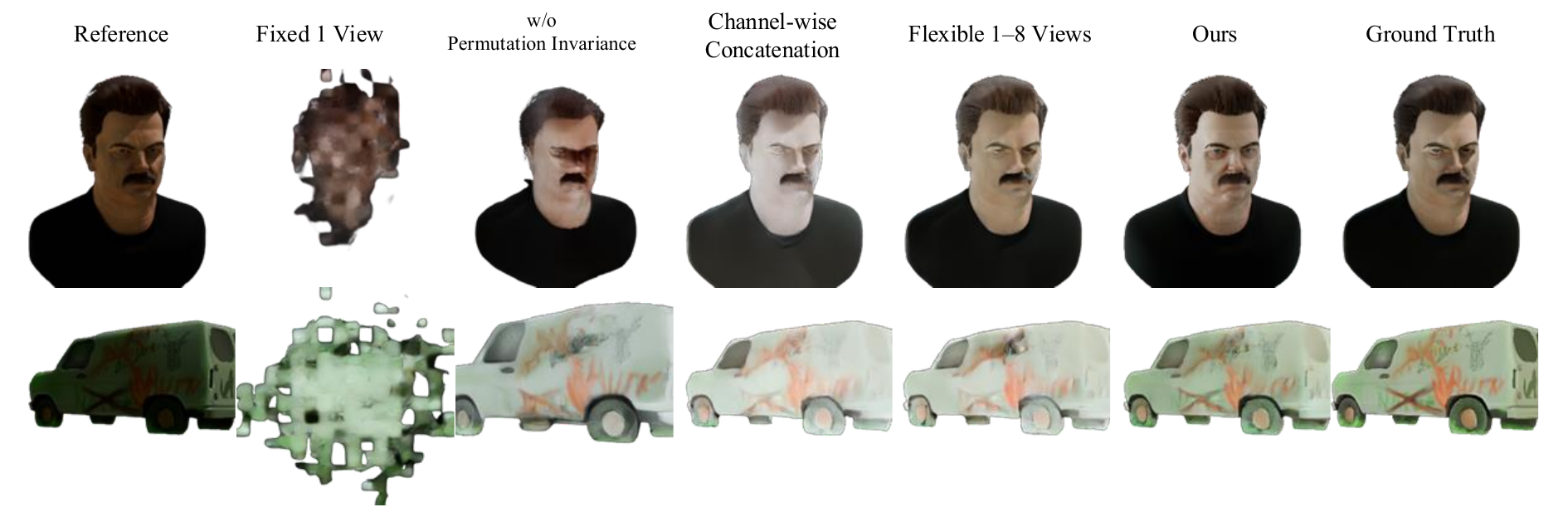} 
\vspace{-20pt}
\caption{Qualitative results of the ablation experiments.}
\label{fig:ablation}
\vspace{-0.1cm}
\end{figure*}

Because training numerous ablated models on the complete training set is computationally prohibitive, we train several variants to convergence on a subset of the training set. We then evaluate them on the full held-out test set, specifically emphasizing their performance in the unseen setting of 32 reference views.

\textbf{1) Ablation on Illumination Conditioning Strategy.} 
A key insight of our work is that prior methods, \textit{e.g.}, Neural Gaffer~\cite{Jin2024} and LightSwitch~\cite{Litman2025}, struggle to effectively inject illumination cues via simple channel-wise concatenation. 
They implicitly assume that the illumination panorama spatially aligns with local image patches. Environment maps encode directional lighting on a spherical domain, which fundamentally differs from image-space pixel coordinates. This domain mismatch hinders effective feature alignment between global illumination and local scene geometry. To validate this observation, we compare two conditioning mechanisms:
\begin{itemize}[leftmargin=*]
\setlength{\itemsep}{1pt}
\setlength{\parsep}{1pt}
\setlength{\parskip}{1pt}
    \item \textbf{Channel-wise Concatenation}: The illumination tokens are directly concatenated with the reference image tokens along the channel dimension.
    \item \textbf{Cross-Attention (Ours)}: Illumination tokens are injected via cross-attention, allowing them to interact with image features.
\end{itemize}
As reported in Table~\ref{tab:analysis_ablation}, the channel-wise concatenation variant suffers from a significant performance drop, highlighting the effectiveness of our cross-attention based latent illumination module. 

\textbf{2) Analysis of Viewpoint Flexibility during Training.} 
We compare three variants that differ in the number of reference views sampled per training iteration:
\begin{itemize}[leftmargin=*]
\setlength{\itemsep}{1pt}
\setlength{\parsep}{1pt}
\setlength{\parskip}{1pt}
    \item \textbf{Fixed Single View:} Exactly one reference image is provided as input in every training iteration.
    \item \textbf{Varying 1--8 Views:} The number of input reference images is uniformly sampled from the range $[1, 8]$.
    \item \textbf{Varying 1--16 Views (Ours):} The number is uniformly sampled from $[1, 16]$ (our default setting).
\end{itemize}
Table~\ref{tab:analysis_ablation} shows that the flexible sampling strategy clearly improves the model's ability to generalize to relighting with up to 32 views.

\textbf{3) Ablation on Permutation-Invariant Design.} 
To enable permutation invariance across input views, we replace the standard Rotary Position Embedding (RoPE)~\cite{rope} used in the vanilla Wan2.1 architecture with PRope~\cite{prope}. 

Table \ref{tab:analysis_ablation} shows a notable performance drop when permutation invariance is absent, highlighting the importance of processing multi-view inputs in an order-agnostic manner. 
More details regarding our analysis and ablations are provided in Appendix ~\ref{app:analysis_and_ablation}.

\begin{table}[t!]\vspace{-0.3cm}
    \centering
    \setlength{\tabcolsep}{3.5pt}
    \caption{Quantitative results of ablation experiments.} \vspace{-0.3cm}
    \label{tab:analysis_ablation}
    \begin{tabular}{lcccc}
        \toprule
        & sPSNR$\uparrow$ & PSNR$\uparrow$ & SSIM$\uparrow$ & LPIPS$\downarrow$ \\
        \midrule
        Channel-wise Concatenation &  23.27 & 20.10 &  0.882 &  0.121 \\
        Fixed Single View & 16.07 &  14.87 &  0.768 &  0.315 \\
        Flexible 1--8 Views &  \underline{22.91} & \underline{20.17} & \underline{0.882} & \underline{0.121} \\
        w/o Permutation Invariance & 17.53 &  15.98 &  0.797 & 0.269 \\
        \textbf{\nickname{} (Ours)} &  \textbf{23.51} & \textbf{20.95} &  \textbf{0.892} &  \textbf{0.111} \\
        \bottomrule
    \end{tabular}
     \vspace{-0.4cm}
\end{table}

\section{Conclusion}
\label{sec:conclusion}

We introduce a large feed-forward model for object relighting from single- and multi-view images, bypassing explicit inverse rendering. Building on a video generation model, our architecture incorporates a latent illumination module that effectively integrates lighting cues into the generation process. We also construct a large-scale dataset featuring diverse objects under a wide range of illumination conditions. Extensive experiments show that \nickname{} clearly outperforms baselines across multiple benchmarks. In the future, we plan to extend our framework to handle complex, scene-level image relighting. Furthermore, as our method is built upon a video foundation model, adapting this pipeline for dynamic video relighting represents a natural and promising direction for future work.

\begin{acks}
This work was supported in part by National Natural Science Foundation of China under Grant 62271431, in part by Research Grants Council of Hong Kong under Grants 15219125 \& 15228626 \& 15225522, in part by Otto Poon Charitable Foundation Smart Cities Research Institute (8-CDCQ), in part by Research Center for Unmanned Autonomous Systems (1-CE3D), and in part by PolyU Kunpeng \& Ascend Technology Innovation Incubation Center, The Hong Kong Polytechnic University.  
\end{acks}

\clearpage
\label{sec:figs}

\clearpage
\bibliographystyle{ACM-Reference-Format}
\bibliography{reference}
\clearpage

\appendix
\label{sec:app}

\setcounter{page}{1}

\onecolumn
\begin{center}
        \Large
        \vspace{0.5em}Supplementary Material \\
        \vspace{1.0em}
\end{center}

\section{More Details in Methodology}
\label{app:method}

\phantom{xx}\textbf{Variational Autoencoder.}
We adopt the pre-trained Variational Autoencoder (VAE)~\cite{vae} from Wan2.1\cite{Wan2025} to project reference images and environment panoramas into a compact latent representation. The VAE parameters remain frozen throughout training to preserve its learned generative priors. 

Although Wan2.1\cite{Wan2025} employs an asymmetric encoding scheme that differentiates the initial frame from subsequent temporal frames to accommodate video compression, our formulation treats all input modalities symmetrically. Consequently, we uniformly apply the first-frame encoding pathway to all reference images and the environment panorama, ensuring consistent latent alignment across inputs.

\textbf{Ray Representation.}
A ray is typically parameterized by its direction $d \in \mathbb{R}^3$ and a point $p \in \mathbb{R}^3$ on its path. In Plücker coordinates, this is represented as a 6D vector $r = (d, m)^\top$, where $m = p \times d \in \mathbb{R}^3$ denotes the ray moment. 

For standard perspective images, the ray corresponding to a pixel coordinate $u \in \mathbb{R}^3$ (in homogeneous form) is derived from the camera intrinsic matrix $K$ and extrinsic transformation $T = [R \mid t]$:
\begin{equation}
    d = R^\top K^{-1} u, \qquad m = c \times d,
\end{equation}
where $c = -R^\top t$ is the camera center expressed in world coordinates and $u$ denotes 2D pixel coordinate.

Environment maps, by contrast, encode global illumination as an HDR equirectangular projection of the spherical radiance field. Unlike perspective images, each pixel corresponds to a collimated ray arriving from infinity toward the scene origin. 
Consequently, the reference point $p$ coincides with the coordinate origin $(0,0,0)^\top$ for all directions, which simplifies the moment to $m = 0$. 
The directional component $d$ uniformly parameterizes the unit sphere $\mathbb{S}^2$, yielding a compact Plücker representation $r = (d, 0_{1\times3})^\top$ for all incident lights.

\textbf{Ray Embedding.} The RaysEmbedding module is a shallow Conv2D-based network designed to project Pl\"ucker ray representations into a structured latent feature space. To effectively capture high-frequency spatial details, the network adopts a Sinusoidal Representation Network architecture~\cite{Sitzmann2020}. Rather than being pretrained in isolation, this embedding network is optimized end-to-end jointly with the entire model. This co-optimization ensures that the latent ray representations dynamically adapt to and align with the downstream task objectives, facilitating seamless gradient flow and holistic feature learning. The detailed architectural specifications of the module are summarized in Table~\ref{tab:rays_embedding}.

\begin{table}[h]
    \centering
    \caption{Architectural details of the \texttt{RaysEmbedding} module. The output dimention is 1536 to match the setting of the Transformer}
    \label{tab:rays_embedding}
    \begin{tabular}{lcccccc}
        \toprule
        \textbf{Layer} & \textbf{Kernel} & \textbf{Stride} & \textbf{Padding} & \textbf{Out Channels} & \textbf{Activation} \\
        \midrule
        Conv2d    & $3 \times 3$  & 2               & 1                & 64                  & Sine \\
         Conv2d     & $3 \times 3$  & 2               & 1                & 128                 & Sine \\
        Conv2d      & $3 \times 3$  & 1               & 1                & 1536   & Sine \\
        MaxPool2d          & $2 \times 2$  & 2               & 0                & 1536   & None \\
        \bottomrule
    \end{tabular}
\end{table}

\section{More Details of Laval Objaverse Dataset}
\label{app:dataset}
Our Laval Objaverse Dataset (LOD), TensoIR~\cite{Jin2023}, StanfordORB~\cite{kuang2023stanfordorb}, relitObjaverse dataset~\cite{Jin2024} and the datasets of LightSwitch~\cite{Litman2025}, and ROGR~\cite{Tang2025} can be used for multi-view relighting. However, because we use entirely open-source materials, our dataset can be fully open-sourced and publicly accessible. Conversely, other datasets are either withheld from public release due to commercial licensing restrictions, or are inherently limited in scale and diversity. As a result, ours will be the only available large-scale, open-source multi-view relighting dataset.

We structured our dataset by defining the following components: 
\begin{enumerate}
    \item \textbf{Object filtering.} We utilize objects from the Objaverse dataset. To exclude meshes with poor geometry or materials, we adopt the object selection criteria from relitObjaverse dataset~\cite{Jin2024}.
    \item \textbf{Lighting selection.} We employ open-source illumination datasets, namely the Laval Indoor HDR Dataset~\cite{laval_indoor} and the Laval Outdoor HDR Dataset~\cite{laval_outdoor}, and rotate each illumination map uniformly 16 times. For each object, we sample 16 environment maps in total, comprising an equal number of indoor and outdoor maps.
    \item \textbf{Rendering protocol.} During rendering, all objects are scaled to a unit size (i.e., the largest dimension of their bounding box is 1) and placed at the origin. The cameras are uniformly distributed on a sphere surrounding the objects, at a distance of 1.8 or 2.2 units from the center, and are oriented to point towards the scene center.
    \item \textbf{Dataset partitioning.} The training, validation, and test sets are split in an 8:1:1 ratio. To prevent data leakage, all camera poses, illumination conditions, and objects used in the training set are strictly excluded from the validation and test sets.
\end{enumerate}
Our model utilizes a total of 39,008 environment maps, which have been strictly divided into training and test sets. The comparison between our Laval Objaverse Dataset and other datasets is illustrated in Table \ref{tab:dataset_comparison}.

\begin{table}[h]
\centering
\caption{Comparison of our Laval Objaverse Dataset (LOD) and others.}
\label{tab:dataset_comparison}
\begin{tabular}{lcccc}
\toprule
\textbf{Method} & \textbf{\# Objects} & \textbf{\# Environment Maps} & \textbf{Synthetic} & \textbf{Publicly Accessible} \\
\hline
TensoIR~\cite{Jin2023}  & 4                         & 11                      & \checkmark & \checkmark \\
StanfordORB~\cite{kuang2023stanfordorb} & 14                       & 7                    & $\times$   & \checkmark \\
LightSwitch~\cite{Litman2025}                & $\sim 100$k               & $\sim 4$k               & \checkmark & $\times$ \\
ROGR~\cite{Tang2025}        & $\sim 90$k                & argumented from 590 maps  & \checkmark & $\times$ \\
relitObjaverse Dataset~\cite{Jin2024} & 90,545 & $\sim 30$k  & \checkmark & $\times$ \\
OLAVTverse~\cite{Zhou2026b} & 765                       & 9930                    & $\times$   & \checkmark \\
\midrule
Laval Objaverse Dataset (ours) & 90,545  & 39,008 & \checkmark   & \checkmark \\
\bottomrule
\end{tabular}
\end{table}

\section{More Details of Experiments}
\label{app:experiment}

\begin{figure*}[!htbp]  
\centering
\includegraphics[width=0.85\linewidth]{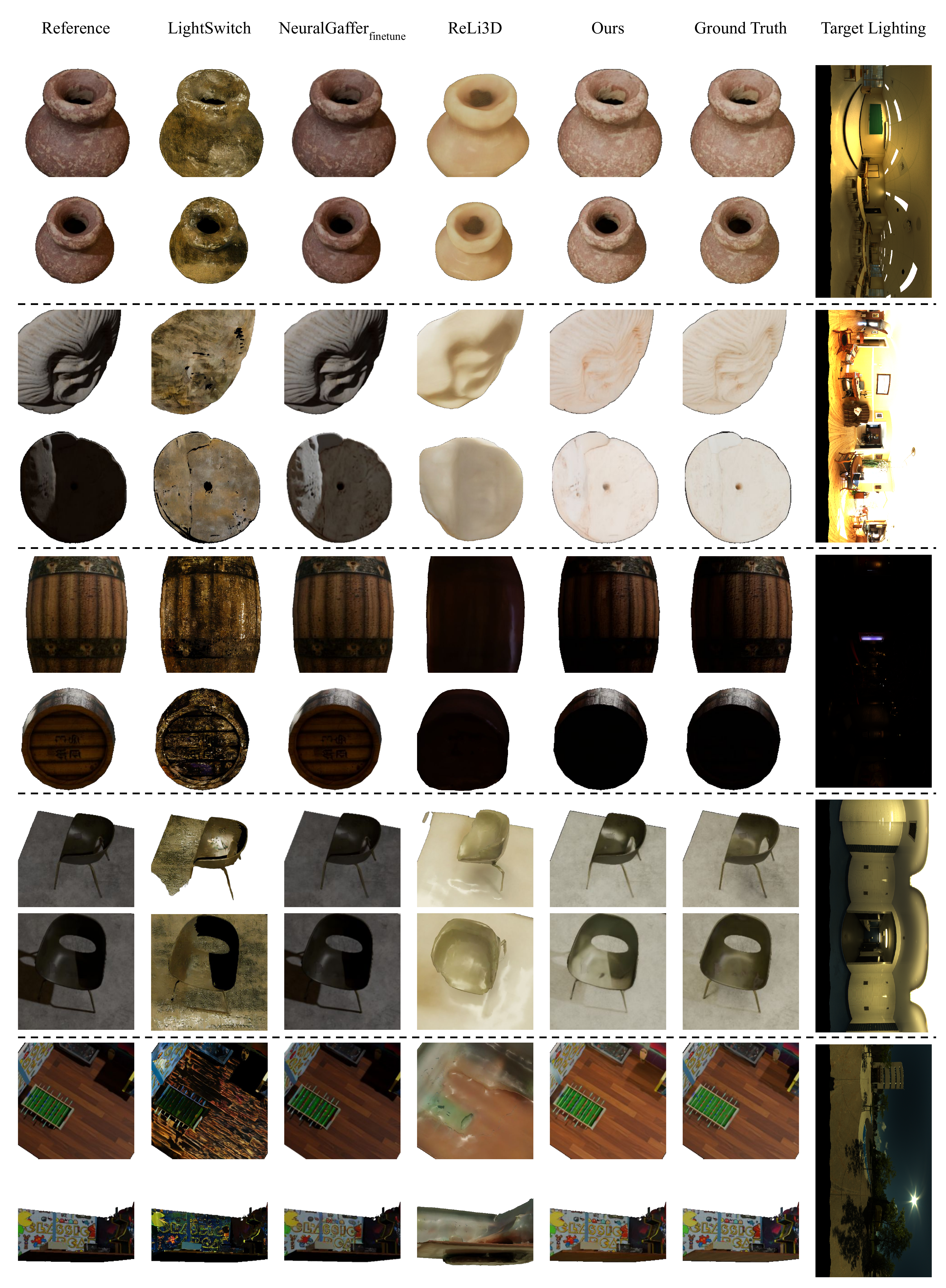} 
\caption{More qualitative results for multi-view image relighting on Laval Objaverse dataset.}
\label{fig:more_qualitaive_of_lavalobjaverse}
\end{figure*}

\begin{figure*}[!htbp]   
\centering
\includegraphics[width=\linewidth]{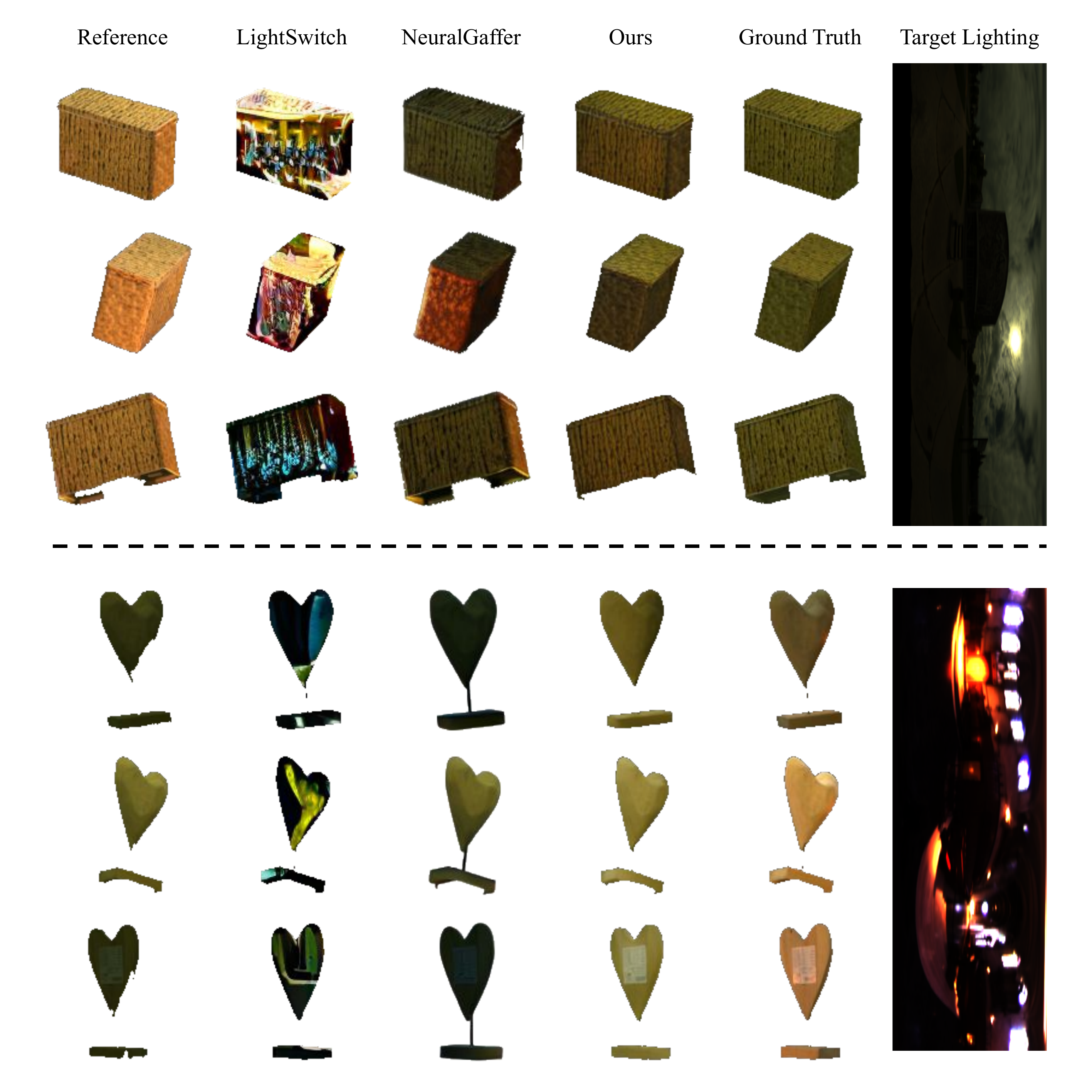} 
\caption{More qualitative results for multi-view image relighting on real-world OLATverse dataset.}
\label{fig:more_qualitaive_of_olatverse}
\end{figure*}

\phantom{xx}\textbf{More results.} Qualitative results on Laval Objaverse dataset and OLATverse dataset~\cite{Zhou2026b} are shown in Figure \ref{fig:more_qualitaive_of_lavalobjaverse} and Figure \ref{fig:more_qualitaive_of_olatverse} respectively. Complete material-stratified quantitative results on the real-world OLATverse dataset~\cite{Zhou2026b} are illustrated in Table \ref{tab:complete_material_stratified}.

\begin{table}[h!]
    \centering
    \caption{Material-stratified quantitative results on the real-world OLATverse dataset~\cite{Zhou2026b}. We report sPSNR$\uparrow$, PSNR$\uparrow$, SSIM$\uparrow$, and LPIPS$\downarrow$.}
    \label{tab:complete_material_stratified}
    \footnotesize
    \setlength{\tabcolsep}{2.5pt}
    \renewcommand{\arraystretch}{1.1}
    \begin{tabular}{l cccc cccc cccc}
        \toprule
        \multirow{2}{*}{\textbf{Method}} & \multicolumn{4}{c}{\textbf{Ceramics}} & \multicolumn{4}{c}{\textbf{Fabric}} & \multicolumn{4}{c}{\textbf{Food}} \\
        \cmidrule(lr){2-5} \cmidrule(lr){6-9} \cmidrule(lr){10-13}
         & sPSNR $\uparrow$ & PSNR $\uparrow$ & SSIM $\uparrow$ & LPIPS $\downarrow$ & sPSNR $\uparrow$ & PSNR $\uparrow$ & SSIM $\uparrow$ & LPIPS $\downarrow$ & sPSNR $\uparrow$ & PSNR $\uparrow$ & SSIM $\uparrow$ & LPIPS $\downarrow$ \\
        \midrule
        DiLightNet~\cite{Zeng2024} & 13.18 & 12.02 & 0.94 & 0.15 & 11.81 & 10.66 & 0.90 & 0.27 & 13.49 & 12.51 & 0.92 & 0.22 \\
        Neural Gaffer~\cite{Jin2024} & 18.24 & 16.38 & \textbf{0.97} & \textbf{0.04} & 17.56 & 14.50 & 0.94 & \underline{0.06} & 19.41 & 17.22 & \textbf{0.96} & \textbf{0.04} \\
        Neural Gaffer$_{finetune}$~\cite{Jin2024} & \underline{19.64} & \underline{16.88} & \textbf{0.97} & \textbf{0.04} & \underline{19.12} & \underline{16.03} & \underline{0.95} & \underline{0.06} & \underline{20.52} & \underline{17.64} & \textbf{0.96} & \textbf{0.04} \\
        LightSwitch~\cite{Litman2025} & 12.70 & 10.10 & 0.93 & 0.13 & 11.17 & 8.90 & 0.90 & 0.16 & 12.73 & 9.98 & 0.92 & 0.14 \\
        \textbf{\nickname{} (Ours)} & \textbf{20.00} & \textbf{17.69} & \textbf{0.97} & \textbf{0.04} & \textbf{20.01} & \textbf{16.61} & \textbf{0.96} & \textbf{0.05} & \textbf{21.18} & \textbf{18.26} & \textbf{0.96} & \textbf{0.04} \\
        \midrule
        \multirow{2}{*}{\textbf{Method}} & \multicolumn{4}{c}{\textbf{Leather}} & \multicolumn{4}{c}{\textbf{Metal}} & \multicolumn{4}{c}{\textbf{Paper}} \\
        \cmidrule(lr){2-5} \cmidrule(lr){6-9} \cmidrule(lr){10-13}
         & sPSNR $\uparrow$ & PSNR $\uparrow$ & SSIM $\uparrow$ & LPIPS $\downarrow$ & sPSNR $\uparrow$ & PSNR $\uparrow$ & SSIM $\uparrow$ & LPIPS $\downarrow$ & sPSNR $\uparrow$ & PSNR $\uparrow$ & SSIM $\uparrow$ & LPIPS $\downarrow$ \\
        \midrule
        DiLightNet~\cite{Zeng2024} & 12.77 & 11.65 & 0.89 & 0.28 & 15.25 & 14.26 & 0.90 & 0.24 & 13.31 & 11.04 & 0.94 & 0.15 \\
        Neural Gaffer~\cite{Jin2024} & 18.79 & 16.15 & 0.95 & \textbf{0.06} & 20.40 & 18.09 & \underline{0.95} & \textbf{0.05} & \underline{18.10} & \underline{15.10} & \textbf{0.97} & \textbf{0.04} \\
        Neural Gaffer$_{finetune}$~\cite{Jin2024} & \underline{20.39} & \underline{17.40} & \textbf{0.96} & \textbf{0.06} & \underline{21.17} & \underline{18.40} & \underline{0.95} & 0.06 & 18.08 & 14.85 & \textbf{0.97} & 0.05 \\
        LightSwitch~\cite{Litman2025} & 12.54 & 9.48 & 0.89 & 0.18 & 14.59 & 10.79 & 0.90 & 0.17 & 10.37 & 8.76 & 0.93 & 0.13 \\
        \textbf{\nickname{} (Ours)} & \textbf{21.09} & \textbf{17.71} & \textbf{0.96} & \textbf{0.06} & \textbf{22.84} & \textbf{19.88} & \textbf{0.96} & \textbf{0.05} & \textbf{19.40} & \textbf{16.16} & \textbf{0.97} & \textbf{0.04} \\
        \midrule
        \multirow{2}{*}{\textbf{Method}} & \multicolumn{4}{c}{\textbf{Plant}} & \multicolumn{4}{c}{\textbf{Plaster}} & \multicolumn{4}{c}{\textbf{Plastic}} \\
        \cmidrule(lr){2-5} \cmidrule(lr){6-9} \cmidrule(lr){10-13}
         & sPSNR $\uparrow$ & PSNR $\uparrow$ & SSIM $\uparrow$ & LPIPS $\downarrow$ & sPSNR $\uparrow$ & PSNR $\uparrow$ & SSIM $\uparrow$ & LPIPS $\downarrow$ & sPSNR $\uparrow$ & PSNR $\uparrow$ & SSIM $\uparrow$ & LPIPS $\downarrow$ \\
        \midrule
        DiLightNet~\cite{Zeng2024} & 12.25 & 12.18 & 0.94 & 0.27 & 12.02 & 11.39 & 0.93 & 0.23 & 9.03 & 8.32 & 0.91 & 0.31 \\
        Neural Gaffer~\cite{Jin2024} & 20.02 & \textbf{18.37} & \textbf{0.98} & \textbf{0.03} & 18.07 & 15.68 & \textbf{0.96} & \textbf{0.05} & 16.92 & 14.19 & \textbf{0.96} & \textbf{0.05} \\
        Neural Gaffer$_{finetune}$~\cite{Jin2024} & \underline{20.69} & 18.34 & \textbf{0.98} & \underline{0.04} & \underline{18.91} & \underline{16.26} & \textbf{0.96} & \textbf{0.05} & \underline{18.59} & \textbf{15.42} & \textbf{0.96} & \textbf{0.05} \\
        LightSwitch~\cite{Litman2025} & 14.26 & 11.52 & 0.95 & 0.11 & 12.21 & 9.87 & 0.93 & 0.13 & 9.83 & 8.46 & 0.92 & 0.15 \\
        \textbf{\nickname{} (Ours)} & \textbf{21.07} & \underline{18.35} & \textbf{0.98} & 0.05 & \textbf{19.59} & \textbf{16.33} & \textbf{0.96} & \textbf{0.05} & \textbf{18.69} & \underline{15.27} & \textbf{0.96} & \textbf{0.05} \\
        \midrule
        \multirow{2}{*}{\textbf{Method}} & \multicolumn{4}{c}{\textbf{Rubber}} & \multicolumn{4}{c}{\textbf{Stone}} & \multicolumn{4}{c}{\textbf{Wax}} \\
        \cmidrule(lr){2-5} \cmidrule(lr){6-9} \cmidrule(lr){10-13}
         & sPSNR $\uparrow$ & PSNR $\uparrow$ & SSIM $\uparrow$ & LPIPS $\downarrow$ & sPSNR $\uparrow$ & PSNR $\uparrow$ & SSIM $\uparrow$ & LPIPS $\downarrow$ & sPSNR $\uparrow$ & PSNR $\uparrow$ & SSIM $\uparrow$ & LPIPS $\downarrow$ \\
        \midrule
        DiLightNet~\cite{Zeng2024} & 15.00 & 13.99 & 0.95 & 0.13 & 9.61 & 9.38 & 0.87 & 0.40 & 9.81 & 9.50 & 0.92 & 0.31 \\
        Neural Gaffer~\cite{Jin2024} & \underline{21.66} & \underline{19.31} & \textbf{0.98} & \textbf{0.03} & 16.51 & 13.90 & \underline{0.94} & \textbf{0.07} & 17.99 & 15.04 & \underline{0.96} & \textbf{0.04} \\
        Neural Gaffer$_{finetune}$~\cite{Jin2024} & 21.25 & 18.56 & \textbf{0.98} & 0.04 & \underline{17.82} & \textbf{15.18} & \underline{0.94} & 0.08 & \underline{19.10} & \underline{15.69} & \underline{0.96} & \textbf{0.04} \\
        LightSwitch~\cite{Litman2025} & 16.08 & 12.07 & 0.96 & 0.10 & 10.99 & 8.42 & 0.89 & 0.18 & 11.25 & 9.71 & 0.93 & 0.13 \\
        \textbf{\nickname{} (Ours)} & \textbf{23.84} & \textbf{20.34} & \textbf{0.98} & \textbf{0.03} & \textbf{18.09} & \underline{15.06} & \textbf{0.95} & \textbf{0.07} & \textbf{19.97} & \textbf{15.78} & \textbf{0.97} & \textbf{0.04} \\
        \midrule
        \multirow{2}{*}{\textbf{Method}} & \multicolumn{4}{c}{\textbf{Wood}} & \multicolumn{4}{c}{\textbf{Furry}} & \multicolumn{4}{c}{\textbf{Glossy}} \\
        \cmidrule(lr){2-5} \cmidrule(lr){6-9} \cmidrule(lr){10-13}
         & sPSNR $\uparrow$ & PSNR $\uparrow$ & SSIM $\uparrow$ & LPIPS $\downarrow$ & sPSNR $\uparrow$ & PSNR $\uparrow$ & SSIM $\uparrow$ & LPIPS $\downarrow$ & sPSNR $\uparrow$ & PSNR $\uparrow$ & SSIM $\uparrow$ & LPIPS $\downarrow$ \\
        \midrule
        DiLightNet~\cite{Zeng2024} & 14.03 & 12.52 & 0.94 & 0.21 & 10.29 & 8.67 & 0.90 & 0.26 & 12.43 & 11.52 & 0.92 & 0.20 \\
        Neural Gaffer~\cite{Jin2024} & 19.34 & 16.67 & \textbf{0.97} & \textbf{0.04} & 15.82 & 12.05 & 0.94 & \underline{0.07} & 17.38 & 15.64 & \textbf{0.96} & \textbf{0.05} \\
        Neural Gaffer$_{finetune}$~\cite{Jin2024} & \underline{20.10} & \underline{16.95} & \textbf{0.97} & 0.05 & \underline{17.43} & \underline{13.99} & \textbf{0.95} & \underline{0.07} & \underline{19.62} & \underline{16.95} & \textbf{0.96} & \textbf{0.05} \\
        LightSwitch~\cite{Litman2025} & 13.15 & 10.51 & 0.93 & 0.13 & 8.98 & 7.89 & 0.89 & 0.17 & 12.41 & 9.83 & 0.92 & 0.14 \\
        \textbf{\nickname{} (Ours)} & \textbf{21.40} & \textbf{17.79} & \textbf{0.97} & \textbf{0.04} & \textbf{17.75} & \textbf{14.28} & \textbf{0.95} & \textbf{0.06} & \textbf{19.91} & \textbf{17.40} & \textbf{0.96} & \textbf{0.05} \\
        \midrule
        \multirow{2}{*}{\textbf{Method}} & \multicolumn{4}{c}{\textbf{Translucent}} & \multicolumn{4}{c}{} & \multicolumn{4}{c}{} \\
        \cmidrule(lr){2-5}
         & sPSNR $\uparrow$ & PSNR $\uparrow$ & SSIM $\uparrow$ & LPIPS $\downarrow$ &  &  &  &  &  &  &  &  \\
        \midrule
        DiLightNet~\cite{Zeng2024} & 8.49 & 8.10 & 0.91 & 0.33 &  &  &  &  &  &  &  &  \\
        Neural Gaffer~\cite{Jin2024} & \underline{17.42} & \textbf{14.06} & \textbf{0.95} & \textbf{0.05} &  &  &  &  &  &  &  &  \\
        Neural Gaffer$_{finetune}$~\cite{Jin2024} & 17.03 & 12.99 & \textbf{0.95} & 0.06 &  &  &  &  &  &  &  &  \\
        LightSwitch~\cite{Litman2025} & 9.74 & 8.65 & 0.92 & 0.16 &  &  &  &  &  &  &  &  \\
        \textbf{\nickname{} (Ours)} & \textbf{18.51} & \underline{13.61} & \textbf{0.95} & \textbf{0.05} &  &  &  &  &  &  &  &  \\
        \bottomrule
    \end{tabular}
\end{table}

\textbf{Testing Data Pairs Curation.}
To ensure a rigorous qualitative assessment of our model and the baselines, we systematically selected 7,248 testing data pairs for each $N$-to-$N$ relighting sub-task ($N \in \{1, 16, 32\}$). 
The pair selection protocol proceeds as follows: for each object in the test split, we randomly designate one illumination as the source and form evaluation pairs by combining this source with each of the 16 available target illuminations (including the source itself). 
For each source--target illumination pair, we randomly sample $N$ viewpoints to constitute the input set. 
This strategy yields 16 test pairs per object, resulting in $453 \times 16 = 7,248$ pairs per sub-task, ensuring comprehensive coverage across diverse lighting and viewing configurations. 
Furthermore, to better simulate real-world deployment conditions, we apply random cropping to each selected image to vary the field of view (FoV), thereby enhancing the representativeness of the test pairs for practical applications.

For environment-map relighting in the OLATverse dataset\cite{Zhou2026b}, we evaluate all 42 objects. For each object, we randomly select 32 viewpoints for assessment. Since OLATverse provides 11 environment maps, we iterate through all lighting combinations, forming 
$11\times11=121$ lighting mappings per object. This results in a total of $42\times 11\times11=5{,082}$
pairs. For material stratification, we utilize the primary material labels annotated by the authors of OLATverse~\cite{Zhou2026b}. For the challenging material categories, \textit{Furry}, \textit{Glossy}, and \textit{Translucent}, we provide the annotations ourselves.

For rotating point-light relighting in the OLATverse dataset\cite{Zhou2026b}, we evaluate all 42 objects using all available viewpoints ($\sim 40$), forming pairs by selecting one point light as the source and another as the target. 

Crucially, all random selections (including viewpoint sampling, illumination assignment, and cropping parameters) are fixed via a predetermined random seed. 
Consequently, every baseline and our method are evaluated on the identical test pairs, guaranteeing a strictly fair and fully reproducible comparison.

\textbf{TensoIR Generalization.}
We validate our model in TensoIR dataset, which is a synthetic dataset and lies out-of-distribution relative to our training data.
For the TensoIR dataset~\cite{Jin2023}, we adapt the protocol accordingly: for each object, we randomly designate one source illumination and iterate over all 14 available target illuminations to form 14 illumination mappings. 
For each mapping, we independently sample 16 viewpoints to construct the 16-image relighting reference set. The TensoIR dataset~\cite{Jin2023} does not include any of the environment maps used during training. The qualitative results are demonstrated on Table~\ref{tab:tensoIR_quantitative}.

\begin{table}[!tb]\vspace{-0.1cm}
\centering
\scriptsize
\setlength{\tabcolsep}{0.pt} 
\caption{Quantitative results for 16 images relighting on TensoIR dataset.}\vspace{-0.2cm}
\label{tab:tensoIR_quantitative}
\begin{tabular}{@{}lcccc@{}}
\toprule
 & \textbf{sPSNR$\uparrow$} & \textbf{PSNR$\uparrow$} & \textbf{SSIM$\uparrow$} & \textbf{LPIPS$\downarrow$} \\
\midrule
DiLightNet~\cite{Zeng2024} & 13.12 & 12.54 & 0.815 & 0.191 \\
Neural Gaffer~\cite{Jin2024} & 18.42 & \underline{16.76} & \textbf{0.906} & 0.106 \\
\small{Neural Gaffer$_{finetuned}$}~\cite{Jin2024} & 18.86 & 16.40 & 0.886 & \textbf{0.090} \\
LightSwitch~\cite{Litman2025} & 16.84 & 15.27 & 0.862 & 0.127 \\
Reli3D~\cite{Dihlmann2026} & \underline{19.51} & 15.04 & 0.889 & 0.137 \\
\midrule
\textbf{\nickname{} (Ours)} & \textbf{19.74} & \textbf{17.64} & \underline{0.904} & \underline{0.100} \\
\bottomrule
\end{tabular}\vspace{-0.25cm}
\end{table}

\noindent \textbf{Metrics.}
In our assessment in images relighting, we report four complementary metrics: scaled PSNR (sPSNR), PSNR, SSIM, and LPIPS.
For all metrics, we restrict evaluation to the foreground region: before computation, we apply the object mask to both the rendered output and ground truth, thereby excluding background discrepancies from the quantitative analysis. 
When computing sPSNR, we first rescale the predicted foreground pixel values $x$ to match the dynamic range of the ground truth $y$: The sPSNR is then defined as:
\begin{equation}
    x' = sx,
\end{equation}
where
\begin{equation}
    s = \arg\min_s || sx- y||_2^2,
\end{equation}

For standard PSNR and sPSNR, we compute the MSE solely within the foreground region.
Consistently, SSIM and LPIPS are likewise evaluated on the foreground-masked images.

When evaluating results on Stanford-ORB\cite{kuang2023stanfordorb}, all of the assessment metrics follow the setting of the benchmark and utilize their official code for assessment.
\\

\textbf{Implementation for Neural Gaffer.}
We note that the vanilla Neural Gaffer~\cite{Jin2024} was originally trained on images composited with a white background. 
To ensure a fair comparison during inference, we similarly composite all reference images with a white background when evaluating both the original and our fine-tuned variants. 
Since all reported metrics are computed exclusively on the foreground region, this adaptation does not bias the quantitative assessment and, in fact, aligns with its original training distribution, thereby benefiting its performance.

We finetune Neural Gaffer~\cite{Jin2024} using 8 NVIDIA RTX 3090 GPUs with a global batch size of 64 for 20K iterations. 
All other hyperparameters and architectural settings follow the default configuration provided by the original implementation. 
Consistent with the vanilla model, both training and inference of our finetuned version are conducted on images composited with a white background.

\section{More Details in Analysis}
\label{app:analysis_and_ablation}

\phantom{xx}\textbf{Training}
For efficient variant validation, we use a subset of $2{,}989$ objects with 16 randomly sampled illuminations per object and 16 viewpoints. 
This reduces training cost while preserving evaluation representativeness.

All variants are trained to convergence using a learning rate of $1\times 10^{-4}$ with a cosine decay schedule.

\end{document}